\documentclass[10pt,twocolumn,letterpaper]{article}

\usepackage{cvpr}              % To produce the CAMERA-READY version
\definecolor{cvprblue}{rgb}{0.21,0.49,0.74}
\usepackage[pagebackref,breaklinks,colorlinks,allcolors=cvprblue]{hyperref}

\usepackage{amsmath}
\usepackage{amsfonts}
\usepackage{multirow}
\usepackage{amssymb}
\usepackage{pifont}
\usepackage{tcolorbox}
\usepackage{listings}
\usepackage[table]{xcolor} 
\usepackage{makecell}

\def\paperID{10216} % *** Enter the Paper ID here
\def\confName{CVPR}
\def\confYear{2026}

\title{Multi-Scale Gaussian-Language Map for Zero-shot Embodied Navigation and Reasoning}

\author{Sixian Zhang$^{1,2}$, Yiyao Wang$^{1,2}$, Xinhang Song$^{1,2}$, Keming Zhang$^{1,2}$, Zijian Xu$^{1,2}$, Shuqiang Jiang$^{2,3}$\footnotemark[1]\\
\small \textsuperscript{1}State Key Laboratory of AI Safety, Institute of Computing Technology, Chinese Academy of Sciences, Beijing\\  
\small \textsuperscript{2}University of Chinese Academy of Sciences, Beijing,
\small \textsuperscript{3}Institute of Computing Technology, Chinese Academy of Sciences, Beijing\\
{\tt\small \{sixian.zhang, yiyao.wang, xinhang.song, keming.zhang, zijian.xu\}@vipl.ict.ac.cn,}\\
{\tt\small sqjiang@ict.ac.cn}
}

\begin{document}
\maketitle
\renewcommand{\thefootnote}{\fnsymbol{footnote}}{\footnotetext[1]{Corresponding author.}}
\begin{abstract}
Understanding the geometric and semantic structure of environments is essential for embodied navigation and reasoning. 
Existing semantic mapping methods trade off between explicit geometry and multi-scale semantics,
and lack a native interface for large models, thus requiring additional training of feature projection for semantic alignment. 
To this end, we propose the multi-scale Gaussian-Language Map (GLMap), which introduces three key designs: (1) explicit geometry, (2) multi-scale semantics covering both instance and region concepts, and (3) a dual-modality interface where each semantic unit jointly stores a natural language description and a 3D Gaussian representation. The 3D Gaussians enable compact storage and fast rendering of task-relevant images via Gaussian splatting. To enable efficient incremental construction, we further propose a Gaussian Estimator that analytically derives Gaussian parameters from dense point clouds without gradient-based optimization. 
Experiments on ObjectNav, InstNav, and SQA tasks show that GLMap effectively enhances target navigation and contextual reasoning, while remaining compatible with large-model-based methods in a zero-shot manner. The code is available at \url{https://github.com/sx-zhang/GLMap}.

\end{abstract}    
\section{Introduction}
\label{sec:intro}

% 理解并记忆所处环境的语义对于Embodied agent是重要的执行Embodied task，比如ObjectNav（find a ‘chair’）、InstanceNav（find ‘a monitor on top of a desk’）、Situated Question Answering（I am washing my face in the sink in front of the mirror. What is on the right side of the soap dispenser in front of me?），比如ObjectNav中agent需要根据观察到Object来推断目标的可能位置，InstanceNav需要根据观察到的物体的空间上下文准确定位目标。Question Answering中需要对语义更多维度的理解，比如物体的属性，affordance区域，场景空间的上下文，来回答Situated的环境的多维度的问题。因此map观察到的多维度语义层级和spatial geometry对于Embodied agent是重要的

% 在复杂、开放的三维环境中，具身智能体（embodied agent）需要理解并记忆环境中的多层级语义与空间结构，以支持下游任务的决策与推理。例如，在 ObjectNav 中，智能体需要依据已观测到的“椅子”“桌子”等物体与场景先验，推断目标的可能空间分布；在 InstanceNav 中，目标往往带有细粒度的上下文约束（如 “一台在桌子上方的显示器”），要求智能体结合实例级关系进行精准定位；在 Situated Question Answering（SQA）中，问题不仅涉及物体类别，还常常依赖属性、可供性（affordance）区域、局部与全局场景语义、以及空间上下文等多维知识（如“我正对着洗手池与镜子洗脸，肥皂分配器右侧是什么？”）。因此，将可观测的多维语义层级与几何结构进行一致建图与长期记忆，是实现稳健具身导航与推理的关键。

Understanding and memorizing the geometric and semantic structure of an environment is fundamental for embodied agents performing embodied tasks, i.e., Object Navigation (ObjectNav) \cite{Target-driven_navigation,Chaplot_NIPS20,VLFM}, Instance Navigation (InstNav) \cite{PSL_eccv24,unigoal_cvpr25}, and Situated Question Answering (SQA) \cite{SQA3D_iclr23,scenellm_wacv25}. 
% the agent must not only perceive the scene but also reason about multi-level semantic cues and spatial relationships to accomplish the goal.
Specifically, the agent in ObjectNav (`find a chair') needs to infer the likely location of the target object from observed semantic cues. 
In InstanceNav (`find a monitor on top of a desk'), the agent needs to accurately localize the target based on fine-grained spatial dependencies between multiple instances. 
In SQA task, the agent requires further geometric-semantic understanding of instance and region semantics to answer situated questions (`I’ve got soup heating on the stove. What should I open behind me to let in fresh air?'). 
Thus, maintaining a semantic map that records multi-scale semantic hierarchies aligned with spatial geometry is essential for embodied navigation and reasoning.

% 现有的mapping类型如图1所示（Grid map，Topological map，Dense geometric map），对于几何结构来说，图（b）所示的Topological map使用graph的结构来构建地图，位置结构用边（‘next to’）表示，这导致其难以描述精准的位置关系。而对于Grid map（图a）和Dense geometric map（图c）的区别是将语义信息存到grid还是点云中。而其中一些基于语义类别的map，即仅将观测到语义存在grid或者点云中，由于其仅有语义类别，难以应付需要更丰富语义的Embodied 任务如InstanceNav，SQA。alternative 基于feature的map，选择将visual-language 特征（如clip）存在grid或者点云中（这样说有点啰嗦，看有没有什么比较简洁的说法，和上面句子有点重复了），相比于单个语义类别，feature能提取多尺度的语义信息，但却不是大模型友好的，即现有的大模型通常是以image和language直接作为输入的，这些map中的语义想被大模型利用往往需要训练一个projection来将这些feature和模型token做对齐。（这里需要再找个地方插入说，目前LLM VLM模型很强大，现在很多Embodied agent都是基于这些模型做的，不适配他们不太行）。因此总结下来，一个语义map for Embodied agent需要具有以下几个特性 1 显式的几何结构，用于空间定位 2 具有多层级的语义信息，可以回答出需要不同语义信息的任务 3 大模型优化，可以直接给出与任务相关的语义信息对应的自然语言描述和视觉图片。

% 现有映射范式大体可分为三类（见图1）：栅格地图（Grid map）、拓扑地图（Topological map）、与稠密几何地图（Dense geometric map）。拓扑地图以图结构表示可达性或相邻关系（例如“next to”），尽管在全局路径规划上高效，但在表达精确的几何方位与尺度关系时力有未逮。栅格地图与稠密几何地图的差异主要体现在语义承载载体：前者将语义写入二维/体素栅格，后者将语义绑定于点云/网格。进一步地，许多方法仅记录类别级语义（semantic class），虽可支持基本的语义导航，却难以覆盖 InstanceNav 与 SQA 所需的更丰富的语义维度（属性、可供性、复合关系等）。为提升表达力，近期工作倾向于将视觉-语言特征（如 CLIP）存入栅格或点云中，以替代单一类别标签；这类特征型语义图确实捕获了多尺度语义，但它们并非大模型友好：当前 LLM/VLM 多以图像与自然语言为原生接口，欲充分利用特征型语义图，往往还需额外训练投影/对齐模块将稠密特征转化为模型可消费的 token 表示，增添了系统复杂度与部署门槛。考虑到LLM/VLM 已成为具身智能的通用推理引擎，一种面向具身任务的语义地图理应具备：（1）显式几何结构以进行可靠空间定位；（2）多层级语义表示以覆盖从实例到复合语义的多任务需求；（3）对大模型友好，能直接以自然语言与可渲染图像形式提供与任务相关的语义与视觉证据，无需特征对齐训练。
% \cite{MapSurvey_TMLR25}

\begin{figure}[t]
\begin{centering}
\includegraphics[width=0.98\columnwidth]{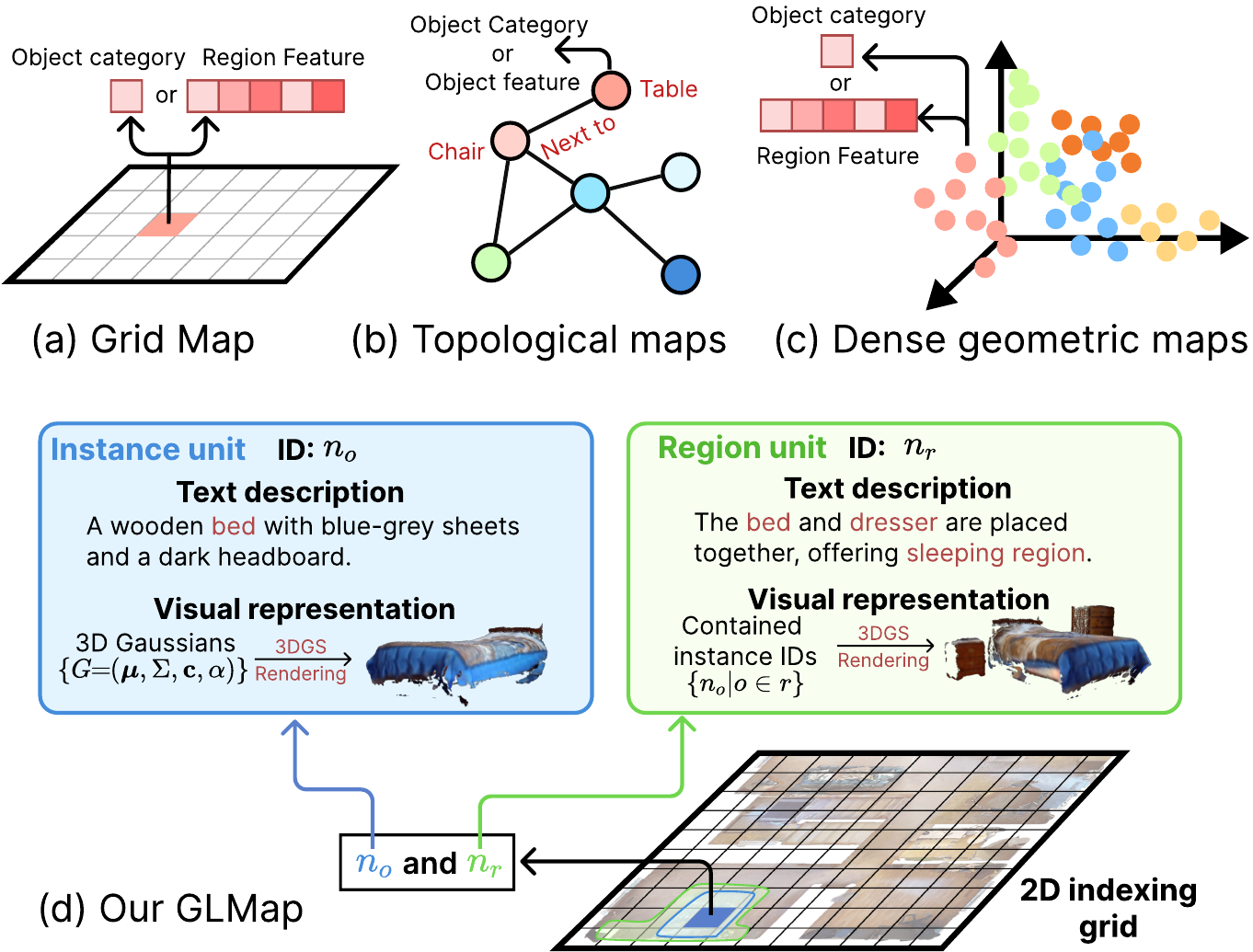}
\end{centering}
\vspace{-4pt}
\caption{\label{fig:into}
Comparison of semantic map structure: (a) Grid map, (b) Topological map, (c) Dense geometric map, and (d) Our GLMap. 
GLMap integrates a 2D indexing grid with multi-scale semantics through instance units and region units, each providing explicit text and visual representations, enabling zero-shot compatibility with current large pretrained models.
}
\vspace{-15pt}
\end{figure}

As shown in Fig.~\ref{fig:into}, existing semantic map structures can be broadly categorized into grid maps \cite{Chaplot_NIPS20}, topological maps \cite{Neural_SLAM,SG-Nav-nips24}, and dense geometric maps \cite{scenellm_wacv25}. 
The topological map (Fig.~\ref{fig:into} b) represents space as a connectivity graph, where nodes denote objects and edges encode adjacency (e.g., next to). 
This structure enables efficient global node-to-node planning, but edges are limited in encoding precise spatial relations.
In contrast, grid-based maps(Fig.~\ref{fig:into} a) and dense geometric maps (Fig.~\ref{fig:into} c) maintain explicit spatial mappings in world coordinates.
They mainly differ in where semantics are stored, either within grids \cite{Chaplot_NIPS20,wzh_ICCV23,Dynam3D_nips25} or point clouds \cite{scenellm_wacv25,3dgsmap_iccv25,axelsson2021semantic_cvpr21}.
Among these approaches, category-based maps assign class labels to grids \cite{Chaplot_NIPS20} or point clouds \cite{axelsson2021semantic_cvpr21}. 
They capture only coarse semantics and fail to represent contextual or affordance information.
In contrast, feature-based maps store vision–language features (e.g., CLIP embeddings) in grids \cite{wzh_ICCV23, Dynam3D_nips25} or point clouds \cite{3dgsmap_iccv25,scenellm_wacv25}, encoding region semantics.
However, they fail to delineate instance boundaries and are not naturally aligned with large models.
Most LLMs, VLMs and MLLMs natively process images and text rather than intermediate feature tensors.
Thus, additional projection \cite{scenellm_wacv25,3dllm-nips23} or alignment \cite{Dynam3D_nips25} is required to bridge features and model tokens.
% , which introduces complexity and training cost.
As recent embodied works \cite{VLFM,unigoal_cvpr25,scenellm_wacv25,3dllm-nips23} increasingly build on large models, this misalignment constrains scalability and modularity.
Therefore, an effective map should satisfy: 
1) \textbf{Explicit geometry}, enabling accurate spatial localization;
2) \textbf{Multi-scale semantics}, capturing both instance and region semantics;
3) \textbf{LMs-friendly interface}, exposing semantics explicitly through natural language and images.

% 为此，我们提出【这里需要起个名字】，如图（d）所示。其具备几个关键的设计点，1 通过使用一个2D的grid做空间索引来 确保语义信息的几何位置的准确 2 多尺度语义信息，语义特征不仅有Instance级别的语义类别，还有多种Instance组合构建的复合语义（比如 affordance 区域，场景语义）等等，确保了【这个方法】能够提供多种层级的语义信息 3 对于每一个Instance级别的语义信息，我们使用3D高斯【需要注意这个3D高斯不是原来的，而是改进一点的，不需要训练确定高斯参数，而是利用深度和相机位置得到点云，直接从点云里估计3D高斯的参数，参数可以随着视角的增多，更新，更新的时候带权重更新】和对应的自然语言描述，其中自然语言可以直接输入大模型无需投影对齐token。而3D高斯可以通过3D高斯泼溅来【给个形容词，体现3DGs的生成速度的快，和高效】为模型提供任务需要的视觉图片。两者都无需对齐，可以直接使用。此外我们提出了一个【我的3D高斯，不是现有的方法，是有改进的】，3D高斯参数不需要在任务过程中反向传播确定参数。【方法的名字】由于只存Instance级别的3D高斯，相比于Dense geometric map的存储更紧凑，并且由于空间索引和语义分开存，可以支持多种检索方式，包括基于位置的信息索引（在【编个坐标】位置中有什么物体？），以及基于语义的索引（在哪里能洗手？）

% (i) 2D空间网格索引：使用轻量的2D栅格作为空间索引面板，精确记录语义锚点的几何位置，同时将语义与几何解耦，既利于空间查询（“在(x, y)附近有什么？”），也便于语义检索（“哪里可以wash hands？”）。
% (ii) 多尺度语义层级：语义分为实例级（instance-level）与复合级（compositional-level）。前者聚焦单实例的身份、属性与几何范围；后者以“实例集合+开放词汇描述”的形式表达场景、可供性与关系语义，支撑跨物体、跨区域的推理。
% (iii) 高斯锚定的跨模态表征：对每个实例级语义，我们存储3D高斯与配对的自然语言描述。为高效构建与增量更新，我们提出OEG（Online Estimated Gaussians）：不通过反向传播优化参数，而是依据深度与相机位姿将mask投影为点云，在线估计高斯参数并随多视角到来加权更新。所得到的3D高斯可通过高效的3D Gaussian splatting快速渲染视图，为VLM提供任务相关的“视觉证据”；同时自然语言描述可直接输入LLM，无需特征投影或对齐头。由于只针对实例存储3D高斯，SAGA-Map相较Dense几何地图更为紧凑，且在空间与语义层面均支持灵活检索。

To this end, we propose the multi-scale Gaussian-Language Map (GLMap) as shown in Fig.~\ref{fig:into} d, which introduces three key designs:
1) 2D indexing grid. A 2D grid enables precise localization of semantic units in metric space.
2) Multi-scale semantics. The semantic units include instance-level concepts (e.g., objects, attributes) and region-level concepts (e.g., functional regions, scenes).
3) Dual-modality interface. Each semantic unit jointly stores a natural language description and a 3D Gaussian representation. 
% 我们选择3D Gaussian来作为视觉表示一是因为其相比点云在存储上更紧凑，并且可以借助Gaussian splatting快速rendering of task-relevant images
We adopt 3D Gaussians as the visual representation because they offer more compact storage than point clouds and enable fast rendering of task-relevant images through Gaussian splatting~\cite{3DGS}.
The dual-modality representation enables GLMap to seamlessly adapt to large models in a zero-shot manner without extra projection training.

The original 3DGS \cite{3DGS,compressed3DGS_cvpr24,Pixel-gs_eccv24} aims to render novel views from multi-view images.
% It initializes a set of 3D Gaussians from a sparse point cloud obtained via structure-from-motion, and refines their parameters through differentiable optimization.
It initializes a set of 3D Gaussians and refines them through differentiable optimization.
This design compensates for the lack of direct depth and camera intrinsics.
% However, in embodied tasks, both depth images and camera poses are available, allowing the reconstruction of high-quality dense point clouds.
% 因此我们propose Gaussian Estimator to directly infers Gaussian parameters  from the dense point cloud
% Gaussian Estimator 通过体素化点云，然后在以每个体素格为中心以其Chebyshev distance为1的邻域内，以计算的方式拟合高斯参数，然后按照曲率的大小来合并初始的高斯参数，确保high-curvature regions保留多的高斯保证边缘清晰 而平滑的地方减少高斯 减少开销。得到的高斯参数在Embodied task执行的过程中递增合并不同视角的。GLMap 使用3D高斯作为视觉信息的高效的表示，其通过计算的方式直接获得高斯参数 without optimization as prior 3DGS methods, which designed for suitable for real-time task
In contrast, embodied tasks provide both depth images and camera intrinsics, enabling the reconstruction of high-quality dense point clouds. 
Therefore, we propose a Gaussian Estimator that directly infers Gaussian parameters from dense point clouds. 
Specifically, the point cloud is voxelized, and for each voxel center, Gaussian parameters are analytically fitted within its Chebyshev neighborhood. 
The initial Gaussian parameters are then merged based on local curvature, preserving more Gaussians in high-curvature regions to maintain sharp boundaries, while reducing them in smoother areas to lower memory cost.
The resulting Gaussian parameters also support incremental updates across different viewpoints.
% Consequently, GLMap leverages 3D Gaussians as an efficient visual representation, directly computing Gaussian parameters without optimization as in prior 3DGS methods~\cite{3dgsmap_iccv25,IGL-Nav_iccv25}, which makes it well-suited for real-time tasks.

Our GLMap is incrementally constructed during embodied tasks.
Given a new RGB-D observation, an MLLM predicts semantic descriptions from the RGB image at both the instance and region levels.
For each instance, GroundingDINO~\cite{GroundingDINO} and MobileSAM~\cite{mobileSAM} extract masks, which are projected into 3D to form point clouds for Gaussian parameter estimation via the proposed Gaussian Estimator.
New instances with high geometric and semantic similarity to existing ones are merged, while others are added as new entries.
Each region is matched and merged using instance overlap and text similarity.
The spatial grid is then updated based on the geometry of instances and regions.

We evaluate GLMap on three representative embodied tasks, including zero-shot ObjectNav \cite{unigoal_cvpr25,VLFM}, InstNav \cite{PSL_eccv24}, and SQA \cite{SQA3D_iclr23}.
Experimental results show that the multi-scale semantics encoded by GLMap effectively support target localization, instance identification, and situated reasoning.
Moreover, GLMap integrates seamlessly into existing LLM-, VLM- and MLLM-based methods in a zero-shot manner, and improves performance across diverse simulators, including HM3D \cite{HM3D_data}, MP3D \cite{MP3D_data} and SQA3D \cite{SQA3D_iclr23}.
% Finally, real-world evaluations further validate the effectiveness of GLMap in practical environments.

% 支持多种查询机制Further, these maps can only be queried using class labels, or in more recent work, using text prompts.

\section{Related Works}

\textbf{Embodied Navigation}.
Early ObjectNav methods relied on geometric-semantic priors learned through reinforcement learning~\cite{Target-driven_navigation, A3C, Chaplot_NIPS20, hoz++PAMI, L-sTDE, zsx_ECCV22} or supervised learning~\cite{PONI,zsx_SGM,T-Diff}. 
Recently, large models have enabled zero-shot ObjectNav methods~\cite{Cows_cvpr23, ZSON, ESC_ICML23} that use them as external priors without task-specific training, which are categorized as LLM-, VLM-, or MLLM-based methods.
LLM-based methods typically use structured textual descriptions of nearby frontier objects~\cite{VoroNav_icml24, ESC_ICML23} or represent the 3D environment as a topological scene graph~\cite{SG-Nav-nips24, graphnav_rss23,unigoal_cvpr25}, which is then used to prompt the LLM to infer likely goal locations.
VLM-based methods employ VLM models~\cite{GLIP_cvpr22} to compute the semantic similarity between current observations and the target~\cite{Cows_cvpr23,GAMap_nips24,VLFM,wzh_g3dlf}, guiding the agent toward high-similarity regions.
MLLM-based methods~\cite{Dynam3D_nips25,Zhong2024TopVNavUT} integrate multimodal inputs such as observations, task instructions, and navigation history into a hidden vector, which is then fed to the MLLM for scene understanding and spatial reasoning.

In InstNav, early methods~\cite{PSL_eccv24,ZSON} use a VLM to project instance text descriptions and visual observations into a shared latent space, and then train a policy network to predict actions. However, these implicit encodings fail to capture fine-grained spatial relations. Recent work~\cite{unigoal_cvpr25} explicitly models spatial relations through a scene graph.

Both ObjectNav and InstNav require agents to model environment semantics, but graph-based maps~\cite{SG-Nav-nips24,unigoal_cvpr25,hoz++PAMI} lack precise spatial and semantic details, semantic maps~\cite{Chaplot_NIPS20,ESC_ICML23} store only object categories without rich descriptions, and implicit encoding maps~\cite{VLFM,wzh_g3dlf,Dynam3D_nips25, wangdynam3d}, though semantically dense, obscure instance boundaries and require extra alignment for MLLMs.
To this end, GLMap provides explicit geometric and multi-scale semantics. 
Each semantic unit includes text descriptions and 3D Gaussian renderings, enabling direct use by LLM, VLM, and MLLM methods without alignment training.

\textbf{Embodied Reasoning}.
% 现有的方法，基于点云，需要projection，我们的不需要，直接出text和image
% previous SQA 工作 以3D点云为输入，将其编码成隐式的3D向量，然后将这些向量和LLM的token语义通过训练进行对齐映射。 LL3DA~\cite{chen2024ll3da} directly 将 point clouds 进行编码，s, while Chat-3D~\cite{wang2023chat} adopts a three-stage training strategy to 编码 point cloud 实现与文本token的对齐. PointLLM~\cite{xu2024pointllm} employs a pretrained 3D encoder, whereas LL3DA~\cite{chen2024ll3da} further introduces a Q-former~\cite{li2023blip} to extract 更富有 semantic 信息的 features. 这些工作的思路是将点云编码成隐含向量，然后再对齐文本语义
% Recent work 直接先3D场景用显式的image进行表征，无需与token的对齐就可以输入到MLLM中进行reasoning。 Video-3D-LLM~\cite{zheng2025video}treats multi-view images as video sequences and integrates them into video-based LMMs, while GPT4Scene~\cite{gpt4scene_25} improves upon this by incorporating  bird’s-eye-view (BEV) representations 和对应的关键帧image作为reference.
% Our GLMap是将3D环境中的信息用显式的Dual-modality的文本和rendered image进行表示。其中我们将3D点云转化成3DGS利用提出的 analytically的Gaussian Estimator，这样根据问题需的视角 去rendered对应的image，显式的文本和image可以直接被MLLM利用来进行reasoning 
% Most existing methods rely on training a semantic alignment model to implicitly encode semantics for SQA. In contrast, we propose a zero-shot approach that directly utilizes the semantic information of the environment for SQA, enhancing environmental interpretability and reducing the cost of semantic alignment. Specifically, we identify and store every object and region in the environment, retaining both textual and visual representations for each semantic unit. This enables direct semantic alignment and reasoning within the LLM.
Previous SQA works take 3D point clouds as input, and encode them into implicit 3D vectors, then align these vectors with LLM tokens through training. 
LL3DA~\cite{chen2024ll3da} directly encodes point clouds, while Chat-3D~\cite{wang2023chat} uses a three-stage training strategy to align point cloud features with text tokens. PointLLM~\cite{xu2024pointllm} employs a pretrained 3D encoder, and LL3DA~\cite{chen2024ll3da} further introduces a Q-former~\cite{li2023blip} to extract more semantically rich features. These methods focus on encoding point clouds into latent vectors and then aligning them with textual semantics.
Recent works explicitly represents 3D scenes with images, allowing direct input into MLLMs without token alignment. 
Video-3D-LLM~\cite{zheng2025video} treats multi-view images as video sequences and integrates them into video-based MLLMs, while GPT4Scene~\cite{gpt4scene_25} employs bird’s-eye-view (BEV) and keyframe images as references.
Our GLMap explicitly represents 3D scenes using dual-modality text and rendered images, which can be directly utilized by MLLMs without additional alignment training.
\section{Approach}
\subsection{Gaussian-Language Map\label{sec:task definition}}
\subsubsection{GLMap definition}
\textbf{Structure.}
Our Gaussian-Language Map (GLMap) is defined as: \(\mathcal{M} = \{ m, \mathcal{S}_{o}, \mathcal{S}_{r} \}\),
where \( m \) denotes the 2D grid used to index semantic units, \( \mathcal{S}_{o}=\{(n_o,o)\} \) is the set of instance semantic units $o$ with its ID $n_o$ and \( \mathcal{S}_{r}=\{(n_r, r)\} \) is the set of region semantic units $r$ with its ID $n_r$.
Each cell of \( m \) stores the IDs of all semantic units whose projections fall within its spatial range, enabling spatial queries during embodied tasks,
and each semantic unit encodes both visual representations and language descriptions. 

An instance unit is defined as
\(o = ( \mathcal{G}, T_{o} )\),
where $T_{o}$ is an open-vocabulary textual description (e.g., object category), 
and $\mathcal{G} = \{ G \!\! = \!\! (\boldsymbol{\mu}, \Sigma, \mathbf{c}, \alpha)\}$ denotes a set of 3D Gaussians. 
Each Gaussian \( G \) is parameterized by its mean \(\boldsymbol{\mu}\), covariance \(\Sigma\), color $\mathbf{c}$ and opacity $ \alpha$. 
We adopt 3D Gaussians as the visual representation because it enables explicit and efficient rendering of visual images from arbitrary viewpoints via 3DGS~\cite{3DGS} while providing higher storage efficiency than dense point clouds.

A region unit is defined as 
\(r = ( \mathcal{I}_r, T_{r} )\),
where \(T_{r}\) is a textual description of the regions (e.g., functional region, scene) 
and \(\mathcal{I}_r = \{n_o|o\in r\}\) denotes the IDs of instance units involved in the region $r$. 
Unlike the instance unit, the region unit does not store 3D Gaussians directly to reduce storage overhead. 
However, the visual image of a region unit can still be readily rendered from the fused 3D Gaussian of all constituent instances.

\textbf{Gaussian estimation.}
% The original 3DGS \cite{3DGS,compressed3DGS_cvpr24,Pixel-gs_eccv24} aims to render novel views from multi-view images.
% It initializes a set of 3D Gaussians from a sparse point cloud obtained via structure-from-motion, and refines their parameters through differentiable optimization.
% This design compensates for the lack of direct depth and pose supervision.
% However, in embodied tasks, both depth images and camera poses are available, allowing the reconstruction of high-quality point clouds.
% Continuing to rely on iterative optimization is inefficient and unsuitable for real-time task.
We propose a Gaussian estimator $f_{GE}$ for GLMap that analytically derives Gaussian primitives from the point cloud $\mathcal{P} = \{ \mathbf{p}_i \}$, formally:
{\small
\begin{equation}
\mathcal{G} = f_{GE}(\mathcal{P}), \quad
\mathcal{G} = \{ G \!\! = \!\! (\boldsymbol{\mu}, \Sigma, \mathbf{c}, \alpha) \}
\label{eq: GE function}
\end{equation}
}
We discretize the point cloud into a regular voxel grid, where each voxel $\mathbf{v}$ holds a subset of points $\mathcal{P}_\mathbf{v}\subseteq\mathcal{P}$.
To prevent isolated Gaussians and ensure spatial overlap among neighboring primitives, each voxel $\mathbf{v}$ gathers points from its local neighborhood for parameter fitting. The point set used by voxel $\mathbf{v}$ for parameter estimation is defined as:
{\small
\begin{equation}
\tilde{\mathcal{P}}_{\mathbf{v}}  = \!\!\!\!\!  \bigcup_{\tilde{\mathbf{v}} \in \mathcal{N}(\mathbf{v})} \!\!\!\!\! \mathcal{P}_{\tilde{\mathbf{v}}}, \quad 
\mathcal{N}(\mathbf{v}) = \{\tilde{\mathbf{v}} \mid \|\tilde{\mathbf{v}} \! - \! \mathbf{v}\|_\infty \le 1\}
\end{equation}
}
where $\|\cdot\|_\infty$ denotes the Chebyshev distance.
Based on $\tilde{\mathcal{P}}_{\mathbf{v}}$, the Gaussian parameters are estimated as:
{\small
\begin{equation}
\boldsymbol{\mu}_\mathbf{v} \! = \!\! \frac{1}{|\tilde{\mathcal{P}}_{\mathbf{v}}|} \!\! \sum_{\mathbf{p}_i \in \tilde{\mathcal{P}}_{\mathbf{v}}} \! \mathbf{p}_i,\;
\Sigma_\mathbf{v} \!= \! \frac{1}{|\tilde{\mathcal{P}}_{\mathbf{v}}|} \!\! \sum_{\mathbf{p}_i \in \tilde{\mathcal{P}}_{\mathbf{v}}} \!\! (\mathbf{p}_i \!-\! \boldsymbol{\mu}_\mathbf{v})(\mathbf{p}_i \!-\! \boldsymbol{\mu}_\mathbf{v})^\top \!\! + \epsilon I
\end{equation}
}
Where $\epsilon$ is a small diagonal regularization term. 
% The RGB color $\mathbf{c}_\mathbf{v}$ and opacity $\alpha_\mathbf{v}$ are estimated as the mean of per-point color and detected confidence score.
The RGB color $\mathbf{c}_\mathbf{v}$ is estimated as the mean of per-point colors, and the opacity $\alpha_\mathbf{v}$ is set to a fixed value (e.g., 0.8).

The voxel-based fitting yields redundant Gaussians in smooth areas, so a merge strategy is further proposed to remove them.
For two candidate Gaussians $G_i$ and $G_j$, their similarity is defined:
{\small
\begin{equation}
D(G_i, G_j) = \| \boldsymbol{\mu}_i - \boldsymbol{\mu}_j \|_2 + \lambda_\Sigma \| \Sigma_i - \Sigma_j \|_F + \lambda_c \| \mathbf{c}_i - \mathbf{c}_j \|_2
\end{equation}
}
where $\|\cdot\|_F$ denotes the Frobenius norm distance, and $\lambda_\Sigma$ and $\lambda_c$ balance shape and color similarity. 
Two Gaussians are merged by:
{\small
\begin{equation}
G_{\text{new}} \! \leftarrow \! G_i \oplus G_j, 
\text{ if } D(G_i, G_j) \!\!<\!\! ( 1 + \tau (\kappa(\Sigma_i) \!+\! \kappa(\Sigma_j)))
\label{eq:merge}
\end{equation}
}
where $\tau$ is a scaling factor and
$\kappa(\Sigma) = \lambda_{\min}(\Sigma) / \operatorname{tr}(\Sigma)$ serves as a curvature proxy computed from the covariance matrix $\Sigma$.
Here, $\lambda_{\min}(\Sigma)$ denotes the smallest eigenvalue corresponding to the surface normal direction, and $\operatorname{tr}(\Sigma)$ is the sum of all eigenvalues.
The curvature-aware threshold preserves more Gaussians in high-curvature regions 
while enforcing stronger merging on flatter areas. 
The $\oplus$ operator averages the parameters of the two 
3D Gaussians (see supplementary for more details).

\begin{figure*}[t]
\begin{centering}
\includegraphics[width=0.99\textwidth]{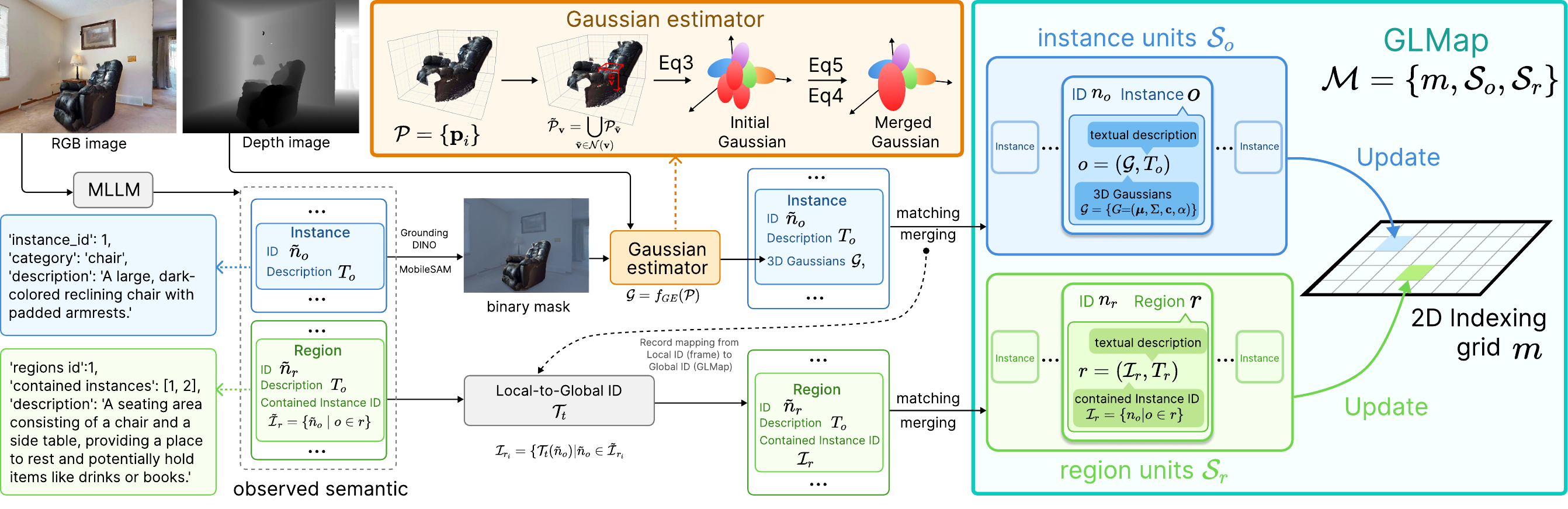}
\par\end{centering}
\vspace{-4pt}
\caption{
\textbf{Incremental update of GLMap.}
The semantics of RGB–D images are first structured into instances and regions. Instance Gaussians are estimated and matched with existing GLMap instances based on textual and Gaussian similarities, and merged accordingly. The matched results determine the global IDs of instances, which are subsequently used for region similarity computation and fusion.
}
\label{fig:framework}
\vspace{-10pt}
\end{figure*}

\subsubsection{Incremental update}
GLMap incrementally maintains consistent semantic and geometric observations during an embodied episode as shown in Fig.~\ref{fig:framework}.
Given the RGB-D image and its corresponding camera pose at timestamp $t$,
The RGB image is first processed by an MLLM for semantic parsing.
The MLLM produces a structured semantic representation defined as:
$\mathcal{S}_t= \{(\tilde{n}_{o}, T_{o})\} \cup \{(\tilde{n}_{r}, T_{r}, \tilde{\mathcal{I}}_{r})\} $
where $\tilde{n}_{o}$ and $\tilde{n}_{r}$ denote the IDs of each instance $o$ and region $r$ in the RGB-D observation. 
$\tilde{\mathcal{I}}_{r} = \{\tilde{n}_{o} \mid o \in r\}$ represents the set of instances contained within region $r$.

Note that two IDs systems are maintained: 
$\tilde{n}$ and $\tilde{\mathcal{I}}_{r}$ are \emph{local IDs} within the current frame, 
while GLMap maintains \emph{global IDs} $n$ and $\mathcal{I}_{r}$. 
The mapping between local and global IDs is determined during the instance update process by deciding whether a newly observed instance should be merged with an existing one. 

\textbf{Instance unit update.}
For each observed instance $o_i$ with its semantic description $(\tilde{n}_{o_i}, T_{o_i})$, 
we first obtain a segmentation mask by localizing the textual description $T_{o_i}$ in the corresponding RGB frame. 
Specifically, we employ GroundingDINO \cite{GroundingDINO} for open-vocabulary region grounding and MobileSAM \cite{mobileSAM} for mask refinement, yielding the binary mask.
The resulting mask is then used to extract the corresponding pixels from the depth image, 
which are back-projected into 3D space using the camera intrinsics and pose, 
producing the associated point cloud $\mathcal{P}_i$.
Each object point cloud $\mathcal{P}_i$ is then analytically transformed into a set of Gaussians by Eq. \ref{eq: GE function}: $\mathcal{G}_i = f_{GE}(\mathcal{P}_i)$ to describes the geometry and appearance of the observed instance.
% where $\mathcal{G}_i \!=\! \{G_{i}\!\!=\!\!(\boldsymbol{\mu}_{i}, \Sigma_{i}, \mathbf{c}_{i}, \alpha_{i})\}$ describes the geometry and appearance of the observed instance.

Each observed instance is then matched to existing instances in the GLMap 
$\mathcal{M} = \{ m, \mathcal{S}_{o}, \mathcal{S}_{r} \}$ 
by jointly evaluating semantic and geometric consistency. 
For an observed instance $o_i$ and an existing instance $o_j \in \mathcal{S}_o$, semantic consistency holds when the cosine similarity of their text description exceeds a predefined threshold, i.e.,
\(\cos\big(\phi(T_{o_i}), \phi(T_{o_j})\big)>\tau_s\), 
where $\phi(\cdot)$ denotes the text embedding function and $\cos(\cdot)$ represents the cosine similarity. 
Once semantic consistency is satisfied, geometric consistency is further evaluated by checking whether any Gaussians in $\mathcal{G}_i$ and $\mathcal{G}_j$ are mergeable, 
according to the criterion defined in Eq. \ref{eq:merge}. 
If the condition holds, the two instances are merged accordingly:
{\small
\begin{equation}
\mathcal{G}_j \leftarrow \mathcal{G}_j \cup \mathcal{G}_i, 
\quad 
T_{o_j} \leftarrow [T_{o_j}; T_{o_i}]
\end{equation}
}
where $\mathcal{G}_j \cup \mathcal{G}_i$ denotes the union of the two Gaussian sets. 
The merged $\mathcal{G}_j$ is subsequently refined by removing redundant Gaussians according to Eq.~\ref{eq:merge}. 
$[T_{o_j}; T_{o_i}]$ represents the concatenation of two sentences with a delimiter. 
When the continuously concatenated text exceeds a predefined length threshold, a lightweight LLM is invoked to merge the sentences. 
This design ensures the efficiency of GLMap updates.
Subsequently, the 2D indexing grid $m$ is updated based on the positional parameters in $\mathcal{G}_j$.
After the merging, the mapping $\mathcal{T}_t$ from local IDs to global IDs is recorded by $\mathcal{T}_t \cup (\tilde{n}_{o_i} \!\!\rightarrow\! n_{o_j})$.
If no matching instance is found, the observed instance $o_i \!=\! (\mathcal{G}_i, T_{o_i})$ is registered in GLMap with a new assigned ID $n_{o_i}$.
Then, the 2D grid indexing $m$ is updated in the same manner, and the coordinate mapping is recorded as $\mathcal{T}_t \cup (\tilde{n}_{o_i} \!\!\rightarrow\! n_{o_i})$.

\textbf{Region unit update.}
For each observed region $\tilde{r}_i = (\tilde{n}_{r_i}, T_{r_i}, \tilde{\mathcal{I}}_{r_i})$, the instance IDs it contains are converted from local to global identifiers using the recorded mapping $\mathcal{T}_t$, formulated as $\mathcal{I}_{r_i}=\{\mathcal{T}_t(\tilde{n}_{o})| \tilde{n}_{o} \in \tilde{\mathcal{I}}_{r_i}\}$.

The region $r_i$ is then matched to an existing region $r_j = (\mathcal{I}_{r_j}, T_{r_j}) \in \mathcal{S}_r$ by jointly evaluating semantic consistency and instance-set consistency.
Similar to instance matching, the semantic consistency between regions is satisfied when \(\cos\big(\phi(T_{r_i}), \phi(T_{r_j})\big)>\tau_s\). 
Instance-set consistency holds when the two regions share at least one common instance, i.e., $\mathcal{I}_{r_i} \cap \mathcal{I}_{r_j} \neq \emptyset$.
If both conditions are satisfied, the observed region is merged with the existing one as
{\small
\begin{equation}
\mathcal{I}_{r_j} \leftarrow \mathcal{I}_{r_j} \cup \mathcal{I}_{r_i}, 
\quad 
T_{r_j} \leftarrow [T_{r_j}; T_{r_i}]
\end{equation}
}

The indexing grid $m$ is subsequently updated according to the spatial locations of the instances in the updated $\mathcal{I}_{r_j}$.
If no matching region is found, a new region ID is assigned and the indexing grid $m$ is updated accordingly.

% \subsubsection{Complexity and Memory Analysis}
% For practical textual updates, textual descriptions are temporarily buffered in a stack rather than merged by the LLM at each step. The LLM-based merging is triggered only when the stack reaches its capacity, and a lightweight LLM is employed. These designs jointly ensure efficient GLMap updates.
% 只存Instance，少背景
% 比点云 体素下采样

\subsection{Embodied task with GLMap}
% $\mathcal{M} = \{ m, \mathcal{S}_{o}, \mathcal{S}_{r} \}$ 由于将空间中的信息分成空间位置，Instance语义 region 语义解耦开的结构化存储，因此可以为embodied task进行决策时提供multi-scale的查询，
% 1）可以根据2D indexing grid来进行有关位置的回答，（比如 和位置有关的问题 in SQA ‘What is located at coordinate [6.8, 1.2, 0.5]?’）
% 2）可以提供Instance level的查询，比如对于ObjectNav中根据以观测到的物体类别来推断目标的可能位置，或者根据Instance的
% 3）可以提供region level的语义查询，比如对于ObjectNav中根据区域上的关联来推断目标位置，或者InstanceNav中，根据context信息来推测目标或者精准定位Instance

% 此外由于不同的语义单元中均存有语言描述和视觉image（通过3D高斯泼溅来高效还原图像），因此GLMap可以适配于先有的基于LLM VLM以及MLLM的方法中。对于visual image的获取，由于3DGS的渲染图像需要给定相机位置，对于给定的Instance的3D高斯核，或者将Instance的3D高斯组合成的代表region的 3D高斯核组，为了使得语义信息更丰富，我们选择给定agent的高度下，以平视觉的角度下，能更多投影（不遮挡）3D高斯的方向作为相机位置【给出公式，如何确定相机位姿】，然后在这个视角下还原图像。【给出一个公式，输出是image，输入是3DGS和相机位姿】。下面将介绍GLMap应用于下游任务的做法
Our GLMap $\mathcal{M} = \{ m, \mathcal{S}_{o}, \mathcal{S}_{r} \}$ models the observed environment by decomposing it into a spatial grid, instance semantic units, and region semantic units.
This structured representation enables versatile querying for various embodied tasks.
1) Spatial queries.
The 2D indexing grid $m$ enables spatial queries, e.g., answering questions in SQA such as “I am facing \ldots, What is located directly to my right?”
2) Instance queries.
The instance semantics $\mathcal{S}_{o}$ support object-level reasoning, e.g., inferring potential target locations in ObjectNav based on observed object categories.
3) Region queries.
The region semantics $\mathcal{S}_{r}$ facilitate contextual reasoning, e.g., estimating target locations from region associations in ObjectNav or leveraging contextual cues for instance localization in InstNav.

In addition, since each semantic unit provides both text descriptions and rendered images, GLMap is naturally compatible with existing LLM-, VLM-, and MLLM-based methods. 
Its downstream applications are outlined below.
% For visual image generation, the camera pose is selected to maximize visible projections of the 3D Gaussian set from an egocentric viewpoint at the agent’s height.

% \subsubsection{Zero-shot Embodied Navigation}
In the \textbf{zero-shot ObjectNav} task, the agent starts from a random location within an unseen environment.
In each episode, the agent needs to autonomously explore and navigate toward an object belonging to a user-specified target category (e.g., chair) using real-time RGB-D observations and its sensor pose. 
The agent operates within a discrete action space, consisting of \texttt{move\_forward}, \texttt{turn\_left}, \texttt{turn\_right} and \texttt{stop}. 
An episode is considered successful if the agent issues the \texttt{stop} action within a set distance threshold (e.g., 1$m$) from the target object within a limited number of steps (e.g., 500). 
The zero-shot setting refers to two aspects: 1) target object categories are drawn from an open vocabulary, and 2) the agent performs navigation without any extra task-specific training.

The \textbf{zero-shot InstNav} task differs from zero-shot ObjectNav in that the navigation goal corresponds to a specific object instance rather than an object category. 
Each episode is associated with a unique target described by a language instruction (e.g., “The chair is made of metal and blue in color, placed next to the table.”), which specifies detailed attributes and spatial context. 
Therefore, InstNav requires a finer-grained understanding of environmental semantics, which can be facilitated by leveraging the region semantic units recorded in our GLMap.

In navigation tasks (e.g., ObjectNav and InstNav), the agent needs to locate goals in unseen environments.
Since exhaustive exploration is inefficient, efficient navigation relies on accurately inferring the location of the goals from the observed semantics.
Recent methods leverage pretrained LLMs \cite{SG-Nav-nips24,FBN_iccv25} or VLMs \cite{VLFM,GAMap_nips24} as priors for location inferring, however, effective reasoning also depends on rich and structured semantic cues.
Our GLMap provides multi-scale semantics to guide this reasoning.
At time step $t$, based on the agent’s RGB-D observations and sensor pose, the GLMap is incrementally built as $\mathcal{M}_t = \{ m_t, \mathcal{S}_{o,t}, \mathcal{S}_{r,t} \}$.
Given the goal description $T_g$, we compute the similarity between the goal and each semantic unit in the GLMap as $s_u = f_S(u, T_g \mid u \in \mathcal{S}_o \cup \mathcal{S}_r)$,
where $f_S$ denotes the similarity function, compatible with both LLM-based and VLM-based methods. Specifically, $f_S$ can be implemented either by prompting LLM to estimate the similarity value between $T_g$ and observed instance $T_o$ and region $T_r$, or by using a VLM to measure the similarity between $T_g$ and the rendered visual image.
% Each instance or region is rendered via 3D Gaussian Splatting \cite{3DGS}, with viewpoints chosen at the agent’s height and planar angles to maximize visible (non-occluded) projections of their 3D Gaussians.
% To enrich semantic observations, we sample camera viewpoints at the agent’s height and along planar angles that maximize the visible (non-occluded) projection of the 3D Gaussians.  
Each instance or region is rendered using 3D Gaussian Splatting \cite{3DGS}, from viewpoints at the agent’s height within navigable areas. The planar angles are selected to maximize the visible (non-occluded) projections of their 3D Gaussians.
Given the similarity scores \( s_u \) and the corresponding 2D grid coordinates \( p_u \) of each semantic unit, we compute a value map \( H(l) \):
{\small
\begin{equation}
H(l) =
\frac{1}{Z}
\sum_{v \in m_t}
\big(\!\!\!\! \sum_{u \in \mathcal{S}_o \cup \mathcal{S}_r}
\!\!\!\! s_u\, \delta(v - p_u)\big)
\mathcal{K}_\sigma(l - v)
\label{eq:gaussian_smooth}
\end{equation}
}
where \( \delta(\cdot) \) denotes the Dirac delta function assigning the relevance score to the corresponding grid cell, and \( \mathcal{K}_\sigma(\cdot) \) is a 2D Gaussian kernel with bandwidth \( \sigma \) that enforces spatial smoothness.  
The next waypoint is selected as the frontier~\cite{FBE} closest to the location of the maximum value grid,
then a local navigation policy (e.g., FMM \cite{FMM}) is used to compute the path toward this waypoint. 
The agent executes next action along this trajectory according to its step size.

\begin{figure*}[t]
\begin{centering}
\includegraphics[width=0.95\textwidth]{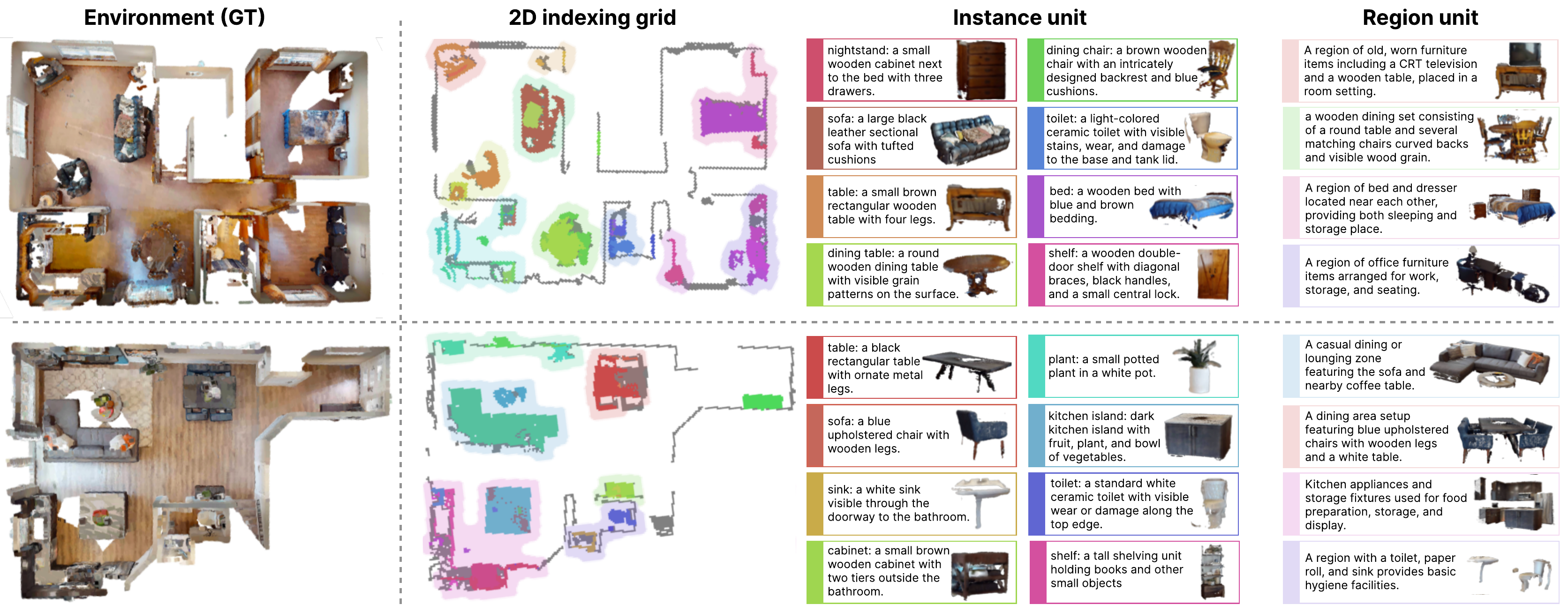}
\par\end{centering}
\vspace{-6pt}
\caption{\textbf{Visualization of GLMap}. The leftmost column shows the 3D ground-truth environment for reference. We visualize three key components of GLMap: the 2D indexing grid, instance unit, and region unit. For each semantic unit, both the recorded textual description and the rendered image produced by 3DGS are shown. Note that only large-volume semantic units are displayed for clarity.
}
\label{fig:vis-GLMap}
\vspace{-15pt}
\end{figure*}

% \subsubsection{Embodied Reasoning}
The \textbf{Situated Question Answering (SQA)} task requires an agent to first understand its situation in the 3D scene from a textual description $T_l$, e.g., ``Sitting at the edge of the bed and facing the couch.'', and then answer a question \( T_q \) under that situation, e.g., ``Can I go straight to the coffee table in front of me?''. 
Our GLMap $\mathcal{M} = \{ m, \mathcal{S}_{o}, \mathcal{S}_{r}\}$ is constructed from egocentric RGB-D video frames together with their camera intrinsics and poses provided by the SQA task.
Then, two spatial probabilities are estimated conditioned on the situation description $T_l$: 
1) the probability that each region \( r \in \mathcal{S}_{r} \) lies near the agent’s situation $p_r=f_{\text{LLM}}(r, T_l)$, and 
2) the probability that each instance \( o \in \mathcal{S}_{o} \) is being faced by the agent $p_o=f_{\text{LLM}}(o, T_l)$.

Both probabilities are projected onto a 2D grid similar to Eq.~\ref{eq:gaussian_smooth}, producing top-down maps \( H_r(l) \) and \( H_o(l) \). 
The agent position and orientation instance are defined as: $l_a = \arg\max_l H_r(l)$ and $l_o = \arg\max_l H_o(l)$.
The orientation is computed by $ \boldsymbol{\theta} = \arctan2(l_o^y - l_a^y,\, l_o^x - l_a^x)$.
Given \( (l_a, \boldsymbol{\theta}) \), we render the visual appearance of the surrounding regions using 3D Gaussian Splatting \cite{3DGS}.
Four camera views are synthesized at the agent’s height, oriented toward $\{\text{front}, \text{back}, \text{left}, \text{right}\}$ relative to $\boldsymbol{\theta}$.
Images rendered from these four views, along with their direction annotations \(\{(I_k, d_k)\}_{k=1}^{4}\), and the question \(T_q\) are jointly fed into a Multimodal Large Language Model (MLLM) to produce the answer: $T_a = f_{\text{MLLM}}\big( \{ (I_k, d_k) \}_{k=1}^{4},\, T_q \big)$.
Since GLMap inherently provides both rendered images and textual representations, it can provide explicit references for MLLMs to perform reasoning without additional training to align tokens and visual representations.

% 引入language-driven map regularization（可选创新点）：
% 使用LLM对融合的语义标签进行语义聚合（例如“chair”和“stool”聚合为“seat”类）。
% 提出affordance field merging策略：通过语义/几何一致性合并affordance区域。

% Instance-aware的 这个是主要的区别，之前的好像直接拿clip特征做，但没ground到Instance上
\section{Experiments}
\subsection{Experimental Setup}
\textbf{Datasets and Metrics.}
For \textbf{zero-shot ObjectNav}, we evaluate our method on the MP3D~\cite{MP3D_data} and HM3D~\cite{HM3D_data} datasets within the Habitat simulator~\cite{Habitat}. 
% 为了 CameraReady 篇幅而压缩内容
% The HM3D validation split contains 2,000 episodes across 20 scenes and 6 object categories, while the MP3D validation split includes 2,195 episodes across 11 scenes and 21 object categories. 
% InstNav
For \textbf{zero-shot InstNav}, we follow the experimental settings of PSL~\cite{PSL_eccv24} and UniGoal~\cite{unigoal_cvpr25}, where the instance goals are described by text. 
Following previous works~\cite{ApexNAV_RAL25,unigoal_cvpr25}, we report Success Rate (SR) and Success weighted by normalized inverse Path Length (SPL) as the evaluation metrics.
% The evaluation set contains 1,000 test episodes with 795 unique instances. We use the same metrics (SR and SPL) as in ObjectNav.
% SQA
For \textbf{SQA}, we use the SQA3D dataset built on ScanNet~\cite{Scannet_cvpr17} scenes. 
% It comprises 20.4k natural language descriptions and 33.4k diverse questions, divided into training, validation, and test splits. 
We adopt the Exact Match (EM-1) metric and the refined Exact Match (EM-R1) metric, following the setup of GPT4Scene~\cite{gpt4scene_25}.

\begin{table}[t]
\setlength{\tabcolsep}{4pt} \renewcommand{\arraystretch}{1.1}

\caption{\label{tab:ablation}Ablation study of multi-scale semantics (instance unit and region unit) of GLMap in HM3D.}
\vspace{-5pt}
\centering{}%
\begin{tabular}{c|ccc|cc}
\hline 
\multirow{2}{*}{{\footnotesize ID}} & \multirow{2}{*}{{\footnotesize Indexing grid}} & \multirow{2}{*}{{\footnotesize Instance unit}} & \multirow{2}{*}{{\footnotesize Region unit}} & \multicolumn{2}{c}{{\footnotesize ObjectNav}}\tabularnewline
 &  &  &  & {\footnotesize SR(\%)} & {\footnotesize SPL(\%)}\tabularnewline
\hline 
{\footnotesize 1} &  &  &  & {\footnotesize 52.5} & {\footnotesize 30.4}\tabularnewline
{\footnotesize 2} & {\footnotesize$\checkmark$} & {\footnotesize$\checkmark$} &  & {\footnotesize 57.4} & {\footnotesize 31.3} \tabularnewline
{\footnotesize 3} & {\footnotesize$\checkmark$} &  & {\footnotesize$\checkmark$} & {\footnotesize 56.2} & {\footnotesize 30.9} \tabularnewline
{\footnotesize 4} & {\footnotesize$\checkmark$} & {\footnotesize$\checkmark$} & {\footnotesize$\checkmark$} & {\footnotesize\textbf{59.1}} & {\footnotesize\textbf{32.2}} \tabularnewline
\hline 
\end{tabular}
\vspace{-10pt}
\end{table}

\textbf{Implementation Details}.
% 在预训练模型使用上，我们选择 Gemma3:27B 作为 MLLM 来首先感知主视图，生成实例和区域描述，然后使用 GroundingDINO 和 MobileSAM 作为检测和分割模型将实例和区域描述进行定位；在 GLMap 的描述合并重生成等文本操作时，我们使用 Qwen3:8B 作为文本处理模型，使用 nomic-embed-text 作为文本嵌入模型。
% 在超参数的使用上，对于合并 Instance 的处理，我们以 embedding 相似度为 0.8 作为语义合并阈值；对于同一个 Instance 的 description，我们在单词数增长超过 200 时执行一次合并重写。对于高斯参数的估计，我们使用 0.01m 作为切分的体素分辨率，对于合并高斯球时使用的 $\lambda_\Sigma$ 和 $\lambda_c$ 参数，我们设置为 $\lambda_\Sigma=0.6$，$\lambda_c=0.4$。
We use the open-source Gemma3-27B~\cite{Kamath2025Gemma3T} as the MLLM to describe instance and region semantics from egocentric RGB images. The generated texts are grounded to obtain masks using GroundingDINO~\cite{GroundingDINO} and MobileSAM~\cite{mobileSAM}.
For Gaussian estimation, the voxel size is set to 1 cm, and the distance weighting hyperparameters are configured as $\lambda_{\Sigma}=0.6$ and $\lambda_{c}=0.4$. 
The text embedding function $\phi(\cdot)$ is implemented with nomic-embed-text~\cite{nussbaum2024nomic}, and the similarity threshold $\tau_s$ is 0.8. 
The buffer length for instance and region text descriptions is 300; once exceeded, a lightweight Qwen3-8B~\cite{qwen3technicalreport} model is invoked for semantic merging.

\begin{figure*}[t]
\begin{centering}
\includegraphics[width=0.95\textwidth]{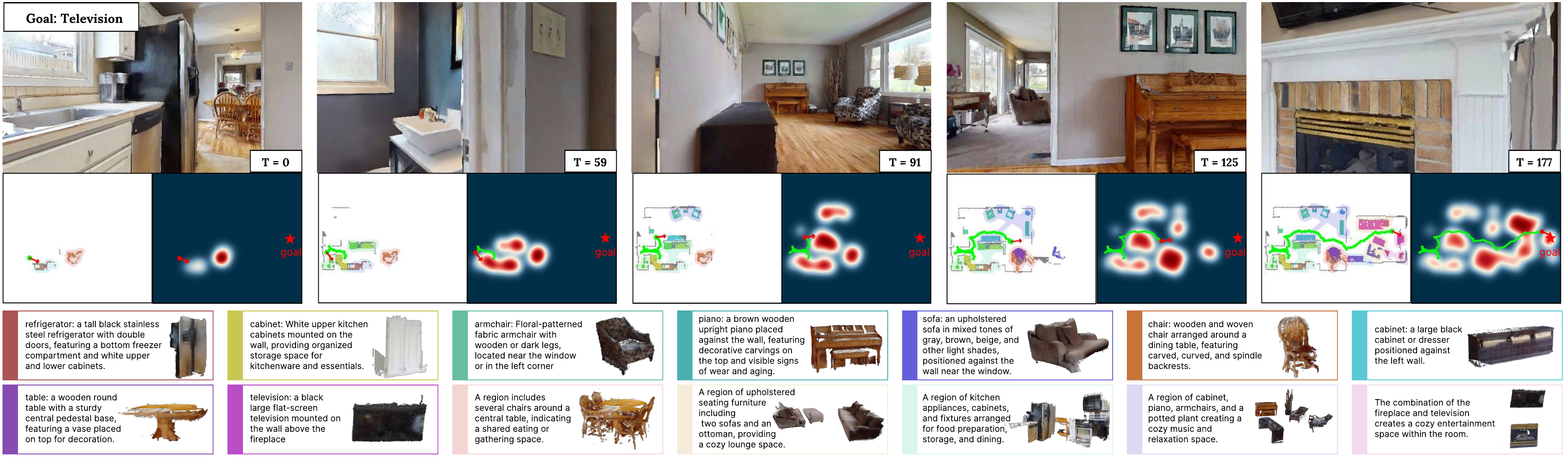}
\par\end{centering}
\vspace{-6pt}
\caption{\textbf{ObjectNav with GLMap}.
% ObjectNav with GLMap，可视化展示了GLMap在导航过程中逐步递增构建，同时展示了根据GLMap中的语义单元计算目标可能出现的value map，可视化展示位置是与真实的坐标是对齐的，尽管导航初期是看不到目标的位置的，此外我们还展示了Instance （靠前的一些）和region unit（靠后的一些）在的语言描述和rendered images。为了展示清晰一些unit被省略了
Although the goal (television) is initially unseen, the value map (computed from semantic units in GLMap) indicates the predicted likelihood of the target’s location, spatially aligned with real-world coordinates.
}
\label{fig:vis-ObjectNav}
\vspace{-10pt}
\end{figure*}

\subsection{Evaluation Results}

\textbf{GLMap Visualization}.
As shown in Fig.~\ref{fig:vis-GLMap}, we visualize the three key components of GLMap: the 2D indexing grid, instance units, and region units. 
By comparing the ground-truth environments with the 2D indexing grids, we observe that the semantic unit localization in GLMap is accurate and spatially consistent. 
For each semantic unit, we present both the recorded textual description and the rendered image generated by our 3D Gaussian estimator. 
Despite deriving Gaussian parameters analytically, the rendered results preserve the original color fidelity and semantic integrity. Moreover, the textual descriptions capture multi-scale semantics, including instance-level concepts (e.g., attribute, and color), as well as region-level concepts (e.g., affordance, spatial relation, and scene).
% 此外 Fig4的可视化，展示了根据GLMap中存储的语义，agent可以有效推测目标的可能位置 in Object Nav
In addition, Fig.~\ref{fig:vis-ObjectNav} shows that the agent can effectively infer goal locations based on the semantics stored in GLMap.

% ablation multi-scale semantic。由于我们的动机是multi-scale semantic是对embodied agent理解环境是重要的，因此我们进行了表1所示的ablation，其中line1对应的baseline是VLFM，它利用VLM计算egocentric view的语义和目标的相似度进行导航。实验结果表明 加入 Instance unit 和region unit都有提升，和到一起还能继续提升。这是因为语义单元中的文本和图像都可以跟目标语义算相似度，能捕捉不同的细节。并且Instance和region的信息互补的 俩还能一起提升

\begin{table}[t]
\setlength{\tabcolsep}{2pt} \renewcommand{\arraystretch}{1}

\caption{\label{tab:LLMs}Evaluations of integrating GLMap into LLM-, VLM-, and MLLM-based methods in a zero-shot manner.}
\vspace{-5pt}
\centering{}%
\begin{tabular}{c|cc|cc}
\hline 
\multirow{2}{*}{{\footnotesize Method}} & \multicolumn{2}{c|}{{\footnotesize ObjectNav}} & \multicolumn{2}{c}{{\footnotesize SQA}}\tabularnewline
 & {\scriptsize SR(\%)} & {\scriptsize SPL(\%)} & {\scriptsize EM-1(\%)} & {\scriptsize EM-R1(\%)}\tabularnewline
\hline 
\rowcolor{gray!15}
\multicolumn{1}{c}{{\scriptsize\textit{LLM-based methods}}} &  & \multicolumn{1}{c}{} &  & \tabularnewline
{\footnotesize ESC \cite{ESC_ICML23}} & {\footnotesize 39.2} & {\footnotesize 22.3} & {\footnotesize -} & {\footnotesize -}\tabularnewline
{\footnotesize ESC+GLMap (Ours)} & {\footnotesize 48.8}{\tiny\textcolor{blue}{(9.6$\uparrow$)}} & {\footnotesize 25.2}{\tiny\textcolor{blue}{(2.9$\uparrow$)}} & {\footnotesize -} & {\footnotesize -}\tabularnewline
\hline 
\rowcolor{gray!15}
\multicolumn{1}{c}{{\scriptsize\textit{VLM-based methods}}} &  & \multicolumn{1}{c}{} &  & \tabularnewline
{\footnotesize VLFM \cite{VLFM}} & {\footnotesize 52.5} & {\footnotesize 30.4} & {\footnotesize -} & {\footnotesize -}\tabularnewline
{\footnotesize VLFM+GLMap (Ours)} & {\footnotesize 59.1}{\tiny\textcolor{blue}{(6.6$\uparrow$)}} & {\footnotesize 32.2}{\tiny\textcolor{blue}{(1.8$\uparrow$)}} & {\footnotesize -} & {\footnotesize -}\tabularnewline
{\footnotesize ApexNAV \cite{ApexNAV_RAL25}} & {\footnotesize 59.6} & {\footnotesize 33.0} & {\footnotesize -} & {\footnotesize -}\tabularnewline
{\footnotesize ApexNAV+GLMap (Ours)} & {\footnotesize 62.7}{\tiny\textcolor{blue}{(3.1$\uparrow$)}} & {\footnotesize 33.7}{\tiny\textcolor{blue}{(0.7$\uparrow$)}} & {\footnotesize -} & {\footnotesize -}\tabularnewline
\hline 
\rowcolor{gray!15}
\multicolumn{1}{c}{{\scriptsize\textit{MLLM-based methods}}} &  & \multicolumn{1}{c}{} &  & \tabularnewline
{\footnotesize GPT4Scene \cite{gpt4scene_25}} & {\footnotesize -} & {\footnotesize -} & {\footnotesize 57.2} & {\footnotesize 60.4}\tabularnewline
{\footnotesize GPT4Scene+GLMap (Ours)} & {\footnotesize -} & {\footnotesize -} & {\footnotesize 58.5}{\tiny\textcolor{blue}{{} (1.3$\uparrow$)}} & {\footnotesize 61.3}{\tiny\textcolor{blue}{(0.9$\uparrow$)}}\tabularnewline
\hline 
\end{tabular}
\vspace{-10pt}
\end{table}

\textbf{Ablation on Multi-scale Semantics}.
Tab.~\ref{tab:ablation} presents the ablation on multi-scale semantics. Line 1 denotes the VLFM~\cite{VLFM} baseline, which navigates by computing semantic similarity between the egocentric view and the target. 
Adding the instance unit and region unit each improves performance, and combining them yields further gains. 
This shows that instance and region semantics offer complementary cues. 
Moreover, compared with directly computing similarity from the egocentric view, the textual and visual descriptions in each semantic unit provide richer cues, enhancing semantic matching and navigation effectiveness.

\begin{table}[t]
\setlength{\tabcolsep}{1pt} \renewcommand{\arraystretch}{1}
\caption{\label{tab:maps}Comparison of GLMap with other mapping structures, including topological, grid, and dense geometric maps, in terms of semantic representation and downstream task performance.}
\vspace{-5pt}
\centering{}%
\begin{tabular}{c|c|ccc}
\hline 
% {\footnotesize Mapping structure} & {\footnotesize Semantic} & {\footnotesize ObjectNav} & {\footnotesize \quad InstNav} & {\footnotesize \quad SQA}\tabularnewline
\multirow{2}{*}{{\footnotesize Mapping structure}} & \multirow{2}{*}{{\footnotesize Semantic unit}} & {\footnotesize ObjectNav} & {\footnotesize InstNav} & {\footnotesize SQA}\tabularnewline
 &  & {\footnotesize SR(\%)} & {\footnotesize SR(\%)} & {\footnotesize EM-1(\%)}\tabularnewline
\hline 
\rowcolor{gray!15}
\multicolumn{1}{c}{{\scriptsize\textit{Topological map}}} & \multicolumn{1}{c}{} &  &  & \tabularnewline
{\footnotesize UniGoal \cite{ESC_ICML23}} & {\scriptsize instance and region (text)} & {\footnotesize 54.5} & {\footnotesize 20.2} & {\footnotesize 34.2}\tabularnewline
\hline 
\rowcolor{gray!15}
\multicolumn{1}{c}{{\scriptsize\textit{Grid map}}} & \multicolumn{1}{c}{} &  &  & \tabularnewline
{\footnotesize GOAT \cite{Goat-bench_cvpr24}} & {\scriptsize object (label)} & {\footnotesize 50.6} & {\footnotesize 17.0} & {\footnotesize -}\tabularnewline
{\footnotesize g3D-LF \cite{wzh_g3dlf}} & {\scriptsize region (visual feature)} & {\footnotesize 55.6} & {\footnotesize 11.5} & {\footnotesize 47.7}\tabularnewline
\hline 
\rowcolor{gray!15}
\multicolumn{1}{c}{{\scriptsize\textit{Dense geometric map}}} & \multicolumn{1}{c}{} &  &  & \tabularnewline
{\footnotesize Chat-Scene\cite{Chat-scene_nips24}} & {\scriptsize region (visual feature)} & {\footnotesize -} & {\footnotesize -} & {\footnotesize 54.6}\tabularnewline
\hline 
% \rowcolor{cyan!10}
% {\raisebox{-.8\height}{\footnotesize GLMap (Ours)}} & \cellcolor{cyan!10}\makecell[ct]{\scriptsize instance and region\\[-0.4ex] \scriptsize (text+rendered image)} & \raisebox{-.8\height}{\footnotesize 62.7} & \raisebox{-.8\height}{\footnotesize 22.5} & \raisebox{-.8\height}{\footnotesize 58.5}\tabularnewline
\rowcolor{cyan!10}
{\footnotesize GLMap (Ours)} &
\shortstack[c]{\scriptsize instance and region\\ \scriptsize (text+rendered image)} &
{\footnotesize \textbf{62.7}} & {\footnotesize \textbf{22.5}} & {\footnotesize \textbf{58.5}}\tabularnewline
% {\footnotesize \raisebox{-.5\height}{GLMap (Ours)}} & 
% \makecell[ct]{\scriptsize instance and region\\[-0.4ex]
% \scriptsize (text+rendered image)} & {\footnotesize 62.7} & {\footnotesize 22.5} & {\footnotesize 58.5}\tabularnewline
\hline 
\end{tabular}
\vspace{-10pt}
\end{table}

\textbf{Adaptability of GLMap to Large Models}.
Since each semantic unit in GLMap jointly stores a text description and a 3D Gaussians, it provides an LM-friendly interface by explicitly exposing both natural language and rendered image for reasoning.
To validate this, we integrate GLMap into several LLM-, VLM-, and MLLM-based methods, as shown in Tab.~\ref{tab:LLMs}.
Across all model categories, GLMap consistently improves performance in a zero-shot setting.
These results show that the text descriptions and 3D Gaussian representations contained in GLMap effectively capture rich semantic details and contextual relationships, and can be seamlessly adapted to diverse model architectures.

% Comparison of GLMap with other mapping structures。我们比较不同map结构如表3所示，其中Topological map，UniGoal 构建 scene graph来描述观察到的语义，其中节点是物体，边是位置关系（如 next to），这种方式对于ObjectNav和InstNav的效果还不错，但对于比较复杂的SQA任务，需要更多的场景细节，但仅文本没法描述全。对于Grid map，其中GOAT采用的是语义地图，每个grid存的是物体类别，由于只有物体类别，ObjectNav还行 但InstNav就比较低了，另外就是这个map的方式没法直接用到SQA上。而另一种做法，g3D-LF是将不同尺度的feature存到grid中，具有比较丰富的语义 因此SQA上还行，但因为不同的feature的获得是以不同patch在主视图上切，对没法描绘清晰的物体轮廓，因此在只要求类别的ObjectNav上还行，但InstNav上就不行了。对于Dense geometric map，Scene-LLM是将CLIP feature存到稀疏化的点云空间，描述了更丰富的语义所以在SQA上性能不错，但他依赖于完整的点云做输入，不支持递增的构建，没法用到ObjectNav和InstNav中。而我们的GLMap，indexing grid记录了显式的位置，比scene graph的位置关系更准，其次对应Instance级别的描述 因此对物体的边界刻画的很清晰相比于feature的，并且其记录了多尺度的语义，因此对于SQA这种需要更丰富语义的，效果也不错

\textbf{Comparison with Other Mapping Structures}.
Tab.~\ref{tab:maps} compares GLMap with topological, grid, and dense geometric maps.
For topological maps, UniGoal~\cite{unigoal_cvpr25} constructs scene graphs where nodes represent objects and edges encode spatial relations such as “next to.” This structure performs well on ObjectNav and InstNav but lacks fine-grained details (e.g., color), limiting its effectiveness on SQA, which requires richer semantic details.
For grid maps, GOAT~\cite{Goat-bench_cvpr24} builds a semantic map where each grid stores only object category labels, supporting basic ObjectNav but performing poorly on InstNav. Moreover, this representation cannot be directly used for MLLMs in SQA. In contrast, g3D-LF~\cite{wzh_g3dlf} encodes visual features into grid cells, providing richer semantics that benefit SQA, though it still requires feature-to-token alignment training. However, since these features are extracted from 2D patches rather than instance-centric regions, InstNav performance declines.
For dense geometric maps, Chat-Scene~\cite{Chat-scene_nips24} encodes 3D point clouds into object-level features and maps them into language tokens for the LLM, achieving strong SQA results. 
However, it relies on complete point clouds, limiting its applicability to incremental tasks like ObjectNav and InstNav.
In contrast, GLMap explicitly indexes positions via a 2D grid and integrates instance and region semantics, enabling accurate localization, clear object boundaries, and multi-scale understanding, which lead to consistent better performances across three tasks.

% TODO: 为了 OA 送审先注释掉以便适配篇幅
% More evaluations (e.g., mathematical formulations, real-world experiments) are detailed in the supplements.

\begin{table}[t]
\setlength{\tabcolsep}{3pt} \renewcommand{\arraystretch}{1.1}

\caption{\label{tab:ObjectNav}Comparison with related methods on \textbf{zero-shot ObjectNav} in MP3D and HM3D.
``TF'' indicates if the method is training-free, and ``OV'' denotes if it supports open-vocabulary object goals.
}
\vspace{-5pt}
\centering{}%
\begin{tabular}{c|cc|cc|cc}
\hline 
\multirow{2}{*}{{\footnotesize Method}} & \multirow{2}{*}{{\footnotesize TF}} & \multirow{2}{*}{{\footnotesize OV}} & \multicolumn{2}{c|}{{\footnotesize MP3D}} & \multicolumn{2}{c}{{\footnotesize HM3D}}\tabularnewline
 &  &  & {\footnotesize SR(\%)} & {\footnotesize SPL(\%)} & {\footnotesize SR(\%)} & {\footnotesize SPL(\%)}\tabularnewline
\hline 
{\footnotesize DD-PPO\cite{DD-PPO}} & {\footnotesize$\times$} & {\footnotesize$\times$} & {\footnotesize 8.0} & {\footnotesize 1.8} & {\footnotesize 27.9} & {\footnotesize 14.2}\tabularnewline
{\footnotesize SemExp \cite{Chaplot_NIPS20}} & {\footnotesize$\times$} & {\footnotesize$\times$} & {\footnotesize 36.0} & {\footnotesize 14.4} & {\footnotesize -} & {\footnotesize -}\tabularnewline
{\footnotesize SGM \cite{zsx_SGM}} & {\footnotesize$\times$} & {\footnotesize$\times$} & {\footnotesize 37.7} & {\footnotesize 14.7} & {\footnotesize 60.2} & {\footnotesize 30.8}\tabularnewline
{\footnotesize T-Diff \cite{T-Diff}} & {\footnotesize$\times$} & {\footnotesize$\times$} & {\footnotesize 39.6} & {\footnotesize 15.2} & - & {\footnotesize -}\tabularnewline
{\footnotesize GOAL \cite{GOAL_nips25}} & {\footnotesize$\times$} & {\footnotesize$\times$} & {\footnotesize 41.7} & {\footnotesize 15.5} & {\footnotesize -} & {\footnotesize -}\tabularnewline
\hline 
{\footnotesize ZSON \cite{ZSON}} & {\footnotesize$\times$} & {\footnotesize$\checkmark$} & {\footnotesize 15.3} & {\footnotesize 4.8} & {\footnotesize 25.5} & {\footnotesize 12.6}\tabularnewline
{\footnotesize PSL \cite{PSL_eccv24}} & {\footnotesize$\times$} & {\footnotesize$\checkmark$} & - & - & {\footnotesize 42.4} & {\footnotesize 19.2}\tabularnewline
\hline 
{\footnotesize ESC \cite{ESC_ICML23}} & {\footnotesize$\checkmark$} & {\footnotesize$\checkmark$} & {\footnotesize 28.7} & {\footnotesize 14.2} & {\footnotesize 39.2} & {\footnotesize 22.3}\tabularnewline
{\footnotesize VLFM \cite{VLFM}} & {\footnotesize$\checkmark$} & {\footnotesize$\checkmark$} & {\footnotesize 36.4} & {\footnotesize 17.5} & {\footnotesize 52.5} & {\footnotesize 30.4}\tabularnewline
{\footnotesize SG-Nav \cite{SG-Nav-nips24}} & {\footnotesize$\checkmark$} & {\footnotesize$\checkmark$} & {\footnotesize 40.2} & {\footnotesize 16.0} & {\footnotesize 54.0} & {\footnotesize 24.9}\tabularnewline
{\footnotesize UniGoal \cite{unigoal_cvpr25}} & {\footnotesize$\checkmark$} & {\footnotesize$\checkmark$} & {\footnotesize 41.0} & {\footnotesize 16.4} & {\footnotesize 54.5} & {\footnotesize 25.1}\tabularnewline
{\footnotesize FBN \cite{FBN_iccv25}} & {\footnotesize$\checkmark$} & {\footnotesize$\checkmark$} & {\footnotesize 41.1} & {\footnotesize 17.3} & {\footnotesize 53.2} & {\footnotesize 30.7}\tabularnewline
{\footnotesize ApexNAV \cite{ApexNAV_RAL25}} & {\footnotesize$\checkmark$} & {\footnotesize$\checkmark$} & {\footnotesize 39.2} & {\footnotesize 17.8} & {\footnotesize 59.6} & {\footnotesize 33.0}\tabularnewline
{\footnotesize BeliefMapNav \cite{beliefmapnav}} & {\footnotesize$\checkmark$} & {\footnotesize$\checkmark$} & {\footnotesize 37.3} & {\footnotesize 17.6} & {\footnotesize 61.4} & {\footnotesize 30.6}\tabularnewline
\hline 
\rowcolor{cyan!10}
{\footnotesize GLMap (Ours)} & {\footnotesize$\checkmark$} & {\footnotesize$\checkmark$} & {\footnotesize\textbf{42.5}} & {\footnotesize\textbf{18.3}} & {\footnotesize\textbf{62.7}} & {\footnotesize\textbf{33.7}}\tabularnewline
\hline 
\end{tabular}
\vspace{-10pt}
\end{table}

\subsection{Comparison with SOTA Methods}

\textbf{Zero-shot ObjectNav}.
We compare three types of ObjectNav methods in Tab.~\ref{tab:ObjectNav}. The first type (DD-PPO~\cite{DD-PPO}, SemExp~\cite{Chaplot_NIPS20}, SGM~\cite{zsx_SGM}, T-Diff~\cite{T-Diff} and GOAL~\cite{GOAL_nips25}) requires task-specific training and operates on a closed set of object goals. 
The second type (ZSON~\cite{ZSON} and PSL~\cite{PSL_eccv24}) also requires training but supports open-vocabulary targets. 
The third type (ESC~\cite{ESC_ICML23}, VLFM~\cite{VLFM}, SG-Nav~\cite{SG-Nav-nips24}, UniGoal~\cite{unigoal_cvpr25}, FBN~\cite{FBN_iccv25}, ApexNAV~\cite{ApexNAV_RAL25} and BeliefMapNav~\cite{beliefmapnav}) is training-free and capable of generalizing to open-vocabulary goals.
Our GLMap provides both text descriptions and rendered images for each instance and region, assisting large models in zero-shot reasoning. 
Since GLMap leverages an open-vocabulary MLLM to generate its initial semantics, it supports open-vocabulary goals. 
Compared with previous methods, GLMap achieves higher performance while remaining training-free, showing the effectiveness of its multi-scale semantics in scene understanding and goal inferring for zero-shot ObjectNav.

\textbf{Zero-shot InstNav}.
We compare GLMap with existing zero-shot InstNav methods in Tab.~\ref{tab:instNav}. ZSON~\cite{ZSON}, PSL~\cite{PSL_eccv24}, and GOAT~\cite{Goat-bench_cvpr24} project the instance text descriptions and visual observations into a shared latent space using a VLM (e.g., CLIP), and train a policy to predict actions from this latent representation. However, such implicit encoding struggles to capture detailed spatial relations, leading to less accurate instance localization. UniGoal~\cite{unigoal_cvpr25} models spatial relations through a scene graph, achieving better localization by explicitly reasoning over instance relationships. 
% However, the textual relation 描述  still 弱于 the visual rendered images of GLMap. 因此 our GLMap 有 in better performance.
However, its textual relational descriptions are less expressive than the visual images rendered in GLMap, which contributes to the improved performance of our method.

\begin{table}[t]
\setlength{\tabcolsep}{8pt} \renewcommand{\arraystretch}{1.1}

\caption{\label{tab:instNav}Comparison with related methods on \textbf{zero-shot InstNav} in HM3D. ``InstRel'' indicates if the method models instance relations (e.g., spatial) to help locate target instances.}
\vspace{-5pt}
\centering{}%
\begin{tabular}{c|c|cc}
\hline 
\multirow{2}{*}{{\footnotesize Method}} & \multirow{2}{*}{{\footnotesize InstRel}} & \multicolumn{2}{c}{{\footnotesize InstNav (HM3D)}}\tabularnewline
 &  & {\footnotesize SR(\%)} & {\footnotesize SPL(\%)}\tabularnewline
\hline 
{\footnotesize ZSON \cite{ZSON}} & {\footnotesize$\times$} & {\footnotesize 10.6} & {\footnotesize 4.9}\tabularnewline
{\footnotesize PSL \cite{PSL_eccv24}} & {\footnotesize$\times$} & {\footnotesize 16.5} & {\footnotesize 7.5}\tabularnewline
{\footnotesize GOAT \cite{Goat-bench_cvpr24}} & {\footnotesize$\times$} & {\footnotesize 17.0} & {\footnotesize 8.8}\tabularnewline
{\footnotesize UniGoal \cite{unigoal_cvpr25}} & {\footnotesize$\checkmark$} & {\footnotesize 20.2} & {\footnotesize 11.4}\tabularnewline
\hline 
\rowcolor{cyan!10}
{\footnotesize GLMap (Ours)} & {\footnotesize$\checkmark$} & {\footnotesize \textbf{22.5}} & {\footnotesize \textbf{13.7}}\tabularnewline
\hline 
\end{tabular}
\vspace{-8pt}
\end{table}

\begin{table}[t]
\setlength{\tabcolsep}{5pt} \renewcommand{\arraystretch}{1.1}

\caption{\label{tab:SQA3D}Comparison with related methods on \textbf{SQA} in SQA3D. ``E-Ref'' indicates if the method provides explicit references for MLLM reasoning rather than implicit embeddings.}
\vspace{-5pt}
\centering{}%
\begin{tabular}{c|c|cc}
\hline 
\multirow{2}{*}{{\footnotesize Method}} & \multirow{2}{*}{{\footnotesize E-Ref}} & \multicolumn{2}{c}{{\footnotesize SQA (SQA3D)}}\tabularnewline
 &  & {\footnotesize EM-1(\%)} & {\footnotesize EM-R1(\%)}\tabularnewline
\hline 
{\footnotesize SQA3D \cite{SQA3D_iclr23}} & {\footnotesize$\times$} & {\footnotesize 46.6} & {\footnotesize -}\tabularnewline
{\footnotesize g3D-LF \cite{wzh_g3dlf}} & {\footnotesize$\times$} & {\footnotesize 47.7} & -\tabularnewline
{\footnotesize Scene-LLM \cite{scenellm_wacv25}} & {\footnotesize$\times$} & {\footnotesize 54.2} & {\footnotesize -}\tabularnewline
{\footnotesize Chat-Scene \cite{Chat-scene_nips24}} & {\footnotesize$\times$} & {\footnotesize 54.6} & {\footnotesize 57.5}\tabularnewline
{\footnotesize 3DGraphLLM \cite{3dgraphllm_iccv25}} & {\footnotesize$\times$} & {\footnotesize 55.9} & - \tabularnewline
{\footnotesize GPT4Scene \cite{gpt4scene_25}} & {\footnotesize$\checkmark$} & {\footnotesize 57.2} & {\footnotesize 60.4}\tabularnewline
\hline 
\rowcolor{cyan!10}
{\footnotesize GLMap (Ours)} & {\footnotesize$\checkmark$} & {\footnotesize \textbf{58.5}} & {\footnotesize \textbf{61.3}}\tabularnewline
\hline 
\end{tabular}
\vspace{-10pt}
\end{table}

% SQA
% 与现有的SQA方法比较如表6，其中SQA3D  Chat-Scene Scene-LLM g3D-LF 是将3D点云或者主视图进行隐式编码，并将编码与MLLM的token对齐使LLM理解3D场景。而3DGraph MLLM是将3D环境编码成一个Graph 然后再训练Graph的编码器for MLLM 帮助MLLM理解环境。这些方法输入到MLLM中的都是隐含式的编码，没有提供显式的reference。GPT4Scene提供给MLLM一个鸟瞰图 并给出关键region的参考图，是一个显式的reference，而我们的GLMap不仅有region的还是Instance的 还有text和image双重信息，从而有更好的性能
\textbf{SQA}.
Tab.~\ref{tab:SQA3D} compares GLMap with existing SQA methods. SQA3D~\cite{SQA3D_iclr23}, g3D-LF~\cite{wzh_g3dlf}, Scene-LLM~\cite{scenellm_wacv25}, and Chat-Scene~\cite{Chat-scene_nips24} implicitly encode 3D scenes from point clouds or egocentric images and align the latent features with MLLM tokens for scene understanding. 
3DGraphLLM~\cite{3dgraphllm_iccv25} represents the 3D environment as a graph and trains a graph encoder to assist MLLM reasoning. 
However, all these methods use implicit embeddings without explicit references. 
In contrast, GPT4Scene offers explicit references by presenting a BEV image together with key region images. Furthermore, GLMap provides explicit references at both the region and instance levels, containing both textual descriptions and rendered visual images, enabling MLLMs to reason more effectively about spatial context and instance relationships, and GLMap achieves better performance.

\section{Conclusion}
In this paper, we propose the multi-scale Gaussian-Language Map (GLMap) for embodied tasks. 
GLMap integrates three main components: 1) explicit geometry, 2) multi-scale semantics covering both instance- and region-level concepts, and 3) a dual-modality interface that jointly encodes a natural language description and a 3D Gaussian representation. 
We also propose a Gaussian Estimator that analytically derives Gaussian parameters from dense point clouds, making the approach suitable for real-time applications.
Experiments on zero-shot ObjectNav, InstNav, and SQA tasks indicate that the multi-scale semantics of GLMap improve embodied understanding, and GLMap complements existing large-model-based methods in a zero-shot setting.

\section*{Acknowledgements}

This work was supported in part by the National Natural Science Foundation of China under Grant 62125207, Grant 62495084, Grant 62272443, and Grant U23B2012, in part by the Beijing Natural Science Foundation under Grant L242020, in part by the Postdoctoral Fellowship Program and China Postdoctoral Science Foundation under Grant Number BX20250391, and in part by the Suzhou Science and Technology Plan Project under grant SYG2024082.

{
    \small
    \bibliographystyle{ieeenat_fullname}
    \bibliography{main}

@String(CVPR= {IEEE Conf. Comput. Vis. Pattern Recog.})

@String(ICCV= {Int. Conf. Comput. Vis.})

@String(ICLR = {Int. Conf. Learn. Represent.})

@String(CVPR  = {CVPR})

@String(ICCV  = {ICCV})

@String(ICLR  = {ICLR})

@inproceedings{A3C,
  author    = {Volodymyr Mnih and
               Adri{\`{a}} Puigdom{\`{e}}nech Badia and
               Mehdi Mirza and
               Alex Graves and
               Timothy P. Lillicrap and
               Tim Harley and
               David Silver and
               Koray Kavukcuoglu},
  title     = {Asynchronous Methods for Deep Reinforcement Learning},
  booktitle = {Proceedings of the 33nd International Conference on Machine Learning,
               {ICML} 2016, New York City, NY, USA, June 19-24, 2016},
  pages     = {1928--1937},
  year      = {2016},
}

@inproceedings{Target-driven_navigation,
  title={Target-driven visual navigation in indoor scenes using deep reinforcement learning},
  author={Zhu, Yuke and Mottaghi, Roozbeh and Kolve, Eric and Lim, Joseph J and Gupta, Abhinav and Fei-Fei, Li and Farhadi, Ali},
  booktitle={2017 IEEE international conference on robotics and automation (ICRA)},
  pages={3357--3364},
  year={2017},
  organization={IEEE}
}

@inproceedings{Habitat,
  author    = {Manolis Savva and
               Jitendra Malik and
               Devi Parikh and
               Dhruv Batra and
               Abhishek Kadian and
               Oleksandr Maksymets and
               Yili Zhao and
               Erik Wijmans and
               Bhavana Jain and
               Julian Straub and
               Jia Liu and
               Vladlen Koltun},
  title     = {Habitat: {A} Platform for Embodied {AI} Research},
  booktitle = {2019 {IEEE/CVF} International Conference on Computer Vision, {ICCV}
               2019, Seoul, Korea (South), October 27 - November 2, 2019},
  pages     = {9338--9346},
  publisher = {{IEEE}},
  year      = {2019},
}

@inproceedings{DD-PPO,
  title={DD-PPO: Learning Near-Perfect PointGoal Navigators from 2.5 Billion Frames},
  author={Wijmans, Erik and Kadian, Abhishek and Morcos, Ari and Lee, Stefan and Essa, Irfan and Parikh, Devi and Savva, Manolis and Batra, Dhruv},
  year={2019},
  booktitle={International Conference on Learning Representations}
}

@article{Chaplot_NIPS20,
  title={Object goal navigation using goal-oriented semantic exploration},
  author={Chaplot, Devendra Singh and Gandhi, Dhiraj Prakashchand and Gupta, Abhinav and Salakhutdinov, Russ R},
  journal={Advances in Neural Information Processing Systems},
  volume={33},
  pages={4247--4258},
  year={2020}
}

@inproceedings{Neural_SLAM,
  title={Neural topological slam for visual navigation},
  author={Chaplot, Devendra Singh and Salakhutdinov, Ruslan and Gupta, Abhinav and Gupta, Saurabh},
  booktitle={Proceedings of the IEEE/CVF conference on computer vision and pattern recognition},
  pages={12875--12884},
  year={2020}
}

@inproceedings{PONI,
  title={Poni: Potential functions for objectgoal navigation with interaction-free learning},
  author={Ramakrishnan, Santhosh Kumar and Chaplot, Devendra Singh and Al-Halah, Ziad and Malik, Jitendra and Grauman, Kristen},
  booktitle={Proceedings of the IEEE/CVF Conference on Computer Vision and Pattern Recognition},
  pages={18890--18900},
  year={2022}
}

@inproceedings{L-sTDE,
  author       = {Sixian Zhang and
                  Xinhang Song and
                  Weijie Li and
                  Yubing Bai and
                  Xinyao Yu and
                  Shuqiang Jiang},
  title        = {Layout-based Causal Inference for Object Navigation},
  booktitle    = {{IEEE/CVF} Conference on Computer Vision and Pattern Recognition,
                  {CVPR} 2023, Vancouver, BC, Canada, June 17-24, 2023},
  pages        = {10792--10802},
  publisher    = {{IEEE}},
  year         = {2023},
}

@inproceedings{FBE,
  author       = {Brian Yamauchi},
  title        = {A frontier-based approach for autonomous exploration},
  booktitle    = {Proceedings 1997 {IEEE} International Symposium on Computational Intelligence
                  in Robotics and Automation CIRA'97 - Towards New Computational Principles
                  for Robotics and Automation, July 10-11, 1997, Monterey, California,
                  {USA}},
  pages        = {146--151},
  publisher    = {{IEEE} Computer Society},
  year         = {1997},
}

@article{FMM,
  title={Fast marching methods},
  author={Sethian, James A},
  journal={SIAM review},
  volume={41},
  number={2},
  pages={199--235},
  year={1999},
  publisher={SIAM}
}

@inproceedings{MP3D_data,
  title={Matterport3D: Learning from RGB-D Data in Indoor Environments},
  author={Chang, Angel and Dai, Angela and Funkhouser, Thomas and Halber, Maciej and Niebner, Matthias and Savva, Manolis and Song, Shuran and Zeng, Andy and Zhang, Yinda},
  booktitle={2017 International Conference on 3D Vision (3DV)},
  pages={667--676},
  year={2017},
  organization={IEEE}
}

@inproceedings{HM3D_data,
  title={Habitat-matterport 3d semantics dataset},
  author={Yadav, Karmesh and Ramrakhya, Ram and Ramakrishnan, Santhosh Kumar and Gervet, Theo and Turner, John and Gokaslan, Aaron and Maestre, Noah and Chang, Angel Xuan and Batra, Dhruv and Savva, Manolis and others},
  booktitle={Proceedings of the IEEE/CVF Conference on Computer Vision and Pattern Recognition},
  pages={4927--4936},
  year={2023}
}

@inproceedings{wzh_ICCV23,
  title={Gridmm: Grid memory map for vision-and-language navigation},
  author={Wang, Zihan and Li, Xiangyang and Yang, Jiahao and Liu, Yeqi and Jiang, Shuqiang},
  booktitle={Proceedings of the IEEE/CVF International Conference on Computer Vision},
  pages={15625--15636},
  year={2023}
}

@inproceedings{zsx_ECCV22,
  title={Generative meta-adversarial network for unseen object navigation},
  author={Zhang, Sixian and Li, Weijie and Song, Xinhang and Bai, Yubing and Jiang, Shuqiang},
  booktitle={European Conference on Computer Vision},
  pages={301--320},
  year={2022},
  organization={Springer}
}

@inproceedings{zsx_SGM,
  title={Imagine before go: Self-supervised generative map for object goal navigation},
  author={Zhang, Sixian and Yu, Xinyao and Song, Xinhang and Wang, Xiaohan and Jiang, Shuqiang},
  booktitle={Proceedings of the IEEE/CVF Conference on Computer Vision and Pattern Recognition},
  pages={16414--16425},
  year={2024}
}

@inproceedings{GroundingDINO,
  title={Grounding DINO: Marrying DINO with Grounded Pre-training for Open-Set Object Detection},
  author={Liu, Shilong and Zeng, Zhaoyang and Ren, Tianhe and Li, Feng and Zhang, Hao and Yang, Jie and Jiang, Qing and Li, Chunyuan and Yang, Jianwei and Su, Hang and others},
  booktitle={European Conference on Computer Vision},
  pages={38--55},
  year={2024}
}

@article{mobileSAM,
  title={Faster segment anything: Towards lightweight sam for mobile applications},
  author={Zhang, Chaoning and Han, Dongshen and Qiao, Yu and Kim, Jung Uk and Bae, Sung-Ho and Lee, Seungkyu and Hong, Choong Seon},
  journal={arXiv preprint arXiv:2306.14289},
  year={2023}
}

@inproceedings{VLFM,
  title={Vlfm: Vision-language frontier maps for zero-shot semantic navigation},
  author={Yokoyama, Naoki and Ha, Sehoon and Batra, Dhruv and Wang, Jiuguang and Bucher, Bernadette},
  booktitle={2024 IEEE International Conference on Robotics and Automation (ICRA)},
  pages={42--48},
  year={2024},
  organization={IEEE}
}

@inproceedings{Cows_cvpr23,
  title={Cows on pasture: Baselines and benchmarks for language-driven zero-shot object navigation},
  author={Gadre, Samir Yitzhak and Wortsman, Mitchell and Ilharco, Gabriel and Schmidt, Ludwig and Song, Shuran},
  booktitle={Proceedings of the IEEE/CVF Conference on Computer Vision and Pattern Recognition},
  pages={23171--23181},
  year={2023}
}

@inproceedings{ESC_ICML23,
  title={Esc: Exploration with soft commonsense constraints for zero-shot object navigation},
  author={Zhou, Kaiwen and Zheng, Kaizhi and Pryor, Connor and Shen, Yilin and Jin, Hongxia and Getoor, Lise and Wang, Xin Eric},
  booktitle={International Conference on Machine Learning},
  pages={42829--42842},
  year={2023},
  organization={PMLR}
}

@article{graphnav_rss23,
  title={How To Not Train Your Dragon: Training-free Embodied Object Goal Navigation with Semantic Frontiers},
  author={Chen, Junting and Li, Guohao and Kumar, Suryansh and Ghanem, Bernard and Yu, Fisher},
  journal={Proceedings of Robotics: Science and System XIX},
  pages={075},
  year={2023},
  publisher={Robotics Science \& Systems Foundation}
}

@inproceedings{GLIP_cvpr22,
  title={Grounded language-image pre-training},
  author={Li, Liunian Harold and Zhang, Pengchuan and Zhang, Haotian and Yang, Jianwei and Li, Chunyuan and Zhong, Yiwu and Wang, Lijuan and Yuan, Lu and Zhang, Lei and Hwang, Jenq-Neng and others},
  booktitle={Proceedings of the IEEE/CVF conference on computer vision and pattern recognition},
  pages={10965--10975},
  year={2022}
}

@article{ZSON,
  title={Zson: Zero-shot object-goal navigation using multimodal goal embeddings},
  author={Majumdar, Arjun and Aggarwal, Gunjan and Devnani, Bhavika and Hoffman, Judy and Batra, Dhruv},
  journal={Advances in Neural Information Processing Systems},
  volume={35},
  pages={32340--32352},
  year={2022}
}

@article{GAMap_nips24,
  title={Gamap: Zero-shot object goal navigation with multi-scale geometric-affordance guidance},
  author={Huang, Hao and Hao, Yu and Wen, Congcong and Tzes, Anthony and Fang, Yi and others},
  journal={Advances in Neural Information Processing Systems},
  volume={37},
  pages={39386--39408},
  year={2024}
}

@inproceedings{VoroNav_icml24,
  title={VoroNav: Voronoi-based Zero-shot Object Navigation with Large Language Model},
  author={Wu, Pengying and Mu, Yao and Wu, Bingxian and Hou, Yi and Ma, Ji and Zhang, Shanghang and Liu, Chang},
  booktitle={International Conference on Machine Learning},
  pages={53757--53775},
  year={2024},
  organization={PMLR}
}

@article{SG-Nav-nips24,
  title={Sg-nav: Online 3d scene graph prompting for llm-based zero-shot object navigation},
  author={Yin, Hang and Xu, Xiuwei and Wu, Zhenyu and Zhou, Jie and Lu, Jiwen},
  journal={Advances in Neural Information Processing Systems},
  volume={37},
  pages={5285--5307},
  year={2024}
}

@article{T-Diff,
  title={Trajectory diffusion for objectgoal navigation},
  author={Yu, Xinyao and Zhang, Sixian and Song, Xinhang and Qin, Xiaorong and Jiang, Shuqiang},
  journal={Advances in Neural Information Processing Systems},
  volume={37},
  pages={110388--110411},
  year={2024}
}

@article{hoz++PAMI,
  title={HOZ++: Versatile Hierarchical Object-to-Zone Graph for Object Navigation},
  author={Zhang, Sixian and Song, Xinhang and Yu, Xinyao and Bai, Yubing and Guo, Xinlong and Li, Weijie and Jiang, Shuqiang},
  journal={IEEE Transactions on Pattern Analysis and Machine Intelligence},
  year={2025},
  publisher={IEEE}
}

@inproceedings{wzh_g3dlf,
  title={g3d-lf: Generalizable 3d-language feature fields for embodied tasks},
  author={Wang, Zihan and Lee, Gim Hee},
  booktitle={Proceedings of the Computer Vision and Pattern Recognition Conference},
  pages={14191--14202},
  year={2025}
}

@article{3DGS,
  title={3D Gaussian splatting for real-time radiance field rendering.},
  author={Kerbl, Bernhard and Kopanas, Georgios and Leimk{\"u}hler, Thomas and Drettakis, George},
  journal={ACM Trans. Graph.},
  volume={42},
  number={4},
  pages={139--1},
  year={2023}
}

@inproceedings{PSL_eccv24,
  title={Prioritized semantic learning for zero-shot instance navigation},
  author={Sun, Xinyu and Liu, Lizhao and Zhi, Hongyan and Qiu, Ronghe and Liang, Junwei},
  booktitle={European Conference on Computer Vision},
  pages={161--178},
  year={2024},
  organization={Springer}
}

@inproceedings{SQA3D_iclr23,
  author       = {Xiaojian Ma and
                  Silong Yong and
                  Zilong Zheng and
                  Qing Li and
                  Yitao Liang and
                  Song{-}Chun Zhu and
                  Siyuan Huang},
  title        = {{SQA3D:} Situated Question Answering in 3D Scenes},
  booktitle    = {The Eleventh International Conference on Learning Representations {ICLR} },
  year         = {2023},
}

@inproceedings{unigoal_cvpr25,
  title={Unigoal: Towards universal zero-shot goal-oriented navigation},
  author={Yin, Hang and Xu, Xiuwei and Zhao, Linqing and Wang, Ziwei and Zhou, Jie and Lu, Jiwen},
  booktitle={Proceedings of the Computer Vision and Pattern Recognition Conference},
  pages={19057--19066},
  year={2025}
}

@article{3dllm-nips23,
  title={3d-llm: Injecting the 3d world into large language models},
  author={Hong, Yining and Zhen, Haoyu and Chen, Peihao and Zheng, Shuhong and Du, Yilun and Chen, Zhenfang and Gan, Chuang},
  journal={Advances in Neural Information Processing Systems},
  volume={36},
  pages={20482--20494},
  year={2023}
}

@article{scenellm_wacv25,
  title={Scene-llm: Extending language model for 3d visual reasoning.},
  author={Fu, Rao and Liu, Jingyu and Chen, Xilun and Nie, Yixin and Xiong, Wenhan},
  journal={IEEE/CVF Winter Conference on Applications of Computer Vision (WACV)},
  volume={2},
  number={3},
  pages={8},
  year={2025}
}

@inproceedings{3dgraphllm_iccv25,
  title={3dgraphllm: Combining semantic graphs and large language models for 3d scene understanding},
  author={Zemskova, Tatiana and Yudin, Dmitry},
  booktitle={Proceedings of the IEEE/CVF International Conference on Computer Vision},
  pages={8885--8895},
  year={2025}
}

@inproceedings{3dgsmap_iccv25,
  title={3d gaussian map with open-set semantic grouping for vision-language navigation},
  author={Gao, Jianzhe and Liu, Rui and Wang, Wenguan},
  booktitle={Proceedings of the IEEE/CVF International Conference on Computer Vision},
  pages={9252--9262},
  year={2025}
}

@inproceedings{axelsson2021semantic_cvpr21,
  title={Semantic labeling of lidar point clouds for uav applications},
  author={Axelsson, Maria and Holmberg, Max and Serra, Sabina and Ovren, Hannes and Tulldahl, Michael},
  booktitle={Proceedings of the IEEE/CVF Conference on Computer Vision and Pattern Recognition},
  pages={4314--4321},
  year={2021}
}

@article{Dynam3D_nips25,
  title={Dynam3D: Dynamic Layered 3D Tokens Empower VLM for Vision-and-Language Navigation},
  author={Wang, Zihan and Lee, Seungjun and Lee, Gim Hee},
  journal={Advances in Neural Information Processing Systems},
  year={2025}
}

@inproceedings{compressed3DGS_cvpr24,
  title={Compressed 3d gaussian splatting for accelerated novel view synthesis},
  author={Niedermayr, Simon and Stumpfegger, Josef and Westermann, R{\"u}diger},
  booktitle={Proceedings of the IEEE/CVF Conference on Computer Vision and Pattern Recognition},
  pages={10349--10358},
  year={2024}
}

@inproceedings{Pixel-gs_eccv24,
  title={Pixel-gs: Density control with pixel-aware gradient for 3d gaussian splatting},
  author={Zhang, Zheng and Hu, Wenbo and Lao, Yixing and He, Tong and Zhao, Hengshuang},
  booktitle={European Conference on Computer Vision},
  pages={326--342},
  year={2024},
}

@inproceedings{FBN_iccv25,
  title={Function-centric Bayesian Network for Zero-Shot Object Goal Navigation},
  author={Zhang, Sixian and Yu, Xinyao and Song, Xinhang and Wang, Yiyao and Jiang, Shuqiang},
  booktitle={Proceedings of the IEEE/CVF International Conference on Computer Vision},
  pages={19535--19545},
  year={2025}
}

@article{ApexNAV_RAL25,
  author       = {Mingjie Zhang and
                  Yuheng Du and
                  Chengkai Wu and
                  Jinni Zhou and
                  Zhenchao Qi and
                  Jun Ma and
                  Boyu Zhou},
  title        = {ApexNAV: An Adaptive Exploration Strategy for Zero-Shot Object Navigation
                  With Target-Centric Semantic Fusion},
  journal      = {{IEEE} Robotics Autom. Lett.},
  volume       = {10},
  number       = {11},
  pages        = {11530--11537},
  year         = {2025},
}

@article{GOAL_nips25,
  author       = {Badi Li and
                  Renjie Lu and
                  Yu Zhou and
                  Jingke Meng and
                  Wei{-}Shi Zheng},
  title        = {Distilling {LLM} Prior to Flow Model for Generalizable Agent's
                  Imagination in Object Goal Navigation},
  journal      = {Advances in Neural Information Processing Systems},
  year         = {2025},
}

@article{beliefmapnav,
  title={BeliefMapNav: 3D Voxel-Based Belief Map for Zero-Shot Object Navigation},
  author={Zhou, Zibo and Hu, Yue and Zhang, Lingkai and Li, Zonglin and Chen, Siheng},
  journal={Advances in Neural Information Processing Systems},
  year={2025},
}

@article{gpt4scene_25,
  title={Gpt4scene: Understand 3d scenes from videos with vision-language models},
  author={Qi, Zhangyang and Zhang, Zhixiong and Fang, Ye and Wang, Jiaqi and Zhao, Hengshuang},
  journal={arXiv preprint arXiv:2501.01428},
  year={2025}
}

@inproceedings{Goat-bench_cvpr24,
  title={Goat-bench: A benchmark for multi-modal lifelong navigation},
  author={Khanna, Mukul and Ramrakhya, Ram and Chhablani, Gunjan and Yenamandra, Sriram and Gervet, Theophile and Chang, Matthew and Kira, Zsolt and Chaplot, Devendra Singh and Batra, Dhruv and Mottaghi, Roozbeh},
  booktitle={Proceedings of the IEEE/CVF Conference on Computer Vision and Pattern Recognition},
  pages={16373--16383},
  year={2024}
}

@article{Chat-scene_nips24,
  title={Chat-scene: Bridging 3d scene and large language models with object identifiers},
  author={Huang, Haifeng and Chen, Yilun and Wang, Zehan and Huang, Rongjie and Xu, Runsen and Wang, Tai and Liu, Luping and Cheng, Xize and Zhao, Yang and Pang, Jiangmiao and others},
  journal={Advances in Neural Information Processing Systems},
  volume={37},
  pages={113991--114017},
  year={2024}
}

@inproceedings{Scannet_cvpr17,
  title={Scannet: Richly-annotated 3d reconstructions of indoor scenes},
  author={Dai, Angela and Chang, Angel X and Savva, Manolis and Halber, Maciej and Funkhouser, Thomas and Nie{\ss}ner, Matthias},
  booktitle={Proceedings of the IEEE conference on computer vision and pattern recognition},
  pages={5828--5839},
  year={2017}
}

@article{Zhong2024TopVNavUT,
  title={TopV-Nav: Unlocking the Top-View Spatial Reasoning Potential of MLLM for Zero-shot Object Navigation},
  author={Linqing Zhong and Chen Gao and Zihan Ding and Yue Liao and Si Liu},
  journal={ArXiv},
  year={2024},
  volume={abs/2411.16425},
  url={https://api.semanticscholar.org/CorpusID:274234805}
}

@inproceedings{chen2024ll3da,
  title={Ll3da: Visual interactive instruction tuning for omni-3d understanding reasoning and planning},
  author={Chen, Sijin and Chen, Xin and Zhang, Chi and Li, Mingsheng and Yu, Gang and Fei, Hao and Zhu, Hongyuan and Fan, Jiayuan and Chen, Tao},
  booktitle={Proceedings of the IEEE/CVF conference on computer vision and pattern recognition},
  pages={26428--26438},
  year={2024}
}

@article{wang2023chat,
  title={Chat-3d: Data-efficiently tuning large language model for universal dialogue of 3d scenes},
  author={Wang, Zehan and Huang, Haifeng and Zhao, Yang and Zhang, Ziang and Zhao, Zhou},
  journal={arXiv preprint arXiv:2308.08769},
  year={2023}
}

@inproceedings{xu2024pointllm,
  title={Pointllm: Empowering large language models to understand point clouds},
  author={Xu, Runsen and Wang, Xiaolong and Wang, Tai and Chen, Yilun and Pang, Jiangmiao and Lin, Dahua},
  booktitle={European Conference on Computer Vision},
  pages={131--147},
  year={2024},
  organization={Springer}
}

@inproceedings{zheng2025video,
  title={Video-3d llm: Learni
         ng position-aware video representation for 3d scene understanding},
  author={Zheng, Duo and Huang, Shijia and Wang, Liwei},
  booktitle={Proceedings of the Computer Vision and Pattern Recognition Conference},
  pages={8995--9006},
  year={2025}
}

@inproceedings{li2023blip,
  title={Blip-2: Bootstrapping language-image pre-training with frozen image encoders and large language models},
  author={Li, Junnan and Li, Dongxu and Savarese, Silvio and Hoi, Steven},
  booktitle={International conference on machine learning},
  pages={19730--19742},
  year={2023},
  organization={PMLR}
}

@article{Kamath2025Gemma3T,
  title={Gemma 3 Technical Report},
  author={Gemma Team Aishwarya Kamath and Johan Ferret and Shreya Pathak and Nino Vieillard and Ramona Merhej and Sarah Perrin and Tatiana Matejovicova and Alexandre Ram'e and Morgane Rivi{\`e}re and Louis Rouillard and Thomas Mesnard and Geoffrey Cideron and Jean-Bastien Grill and Sabela Ramos and Edouard Yvinec and Michelle Casbon and Etienne Pot and Ivo Penchev and Gael Liu and Francesco Visin and Kathleen Kenealy and Lucas Beyer and Xiaohai Zhai and Anton Tsitsulin and R{\'o}bert Istvan Busa-Fekete and Alex Feng and Noveen Sachdeva and Benjamin Coleman and Yi Gao and Basil Mustafa and Iain Barr and Emilio Parisotto and David Tian and Matan Eyal and Colin Cherry and Jan-Thorsten Peter and Danila Sinopalnikov and Surya Bhupatiraju and Rishabh Agarwal and Mehran Kazemi and Dan Malkin and Ravin Kumar and David Vilar and Idan Brusilovsky and Jiaming Luo and Andreas Steiner and Abe Friesen and Abhanshu Sharma and Abheesht Sharma and Adi Mayrav Gilady and Adrian Goedeckemeyer and Alaa Saade and Alexander Kolesnikov and Alexei Bendebury and Alvin Abdagic and Amit Vadi and Andr'as Gyorgy and Andr{\'e} Susano Pinto and Anil Das and Ankur Bapna and Antoine Miech and Antoine Yang and Antonia Paterson and Ashish Shenoy and Ayan Chakrabarti and Bilal Piot and Boxi Wu and Bobak Shahriari and Bryce Petrini and Charlie Chen and Charline Le Lan and Christopher A. Choquette-Choo and Cj Carey and Cormac Brick and Daniel Deutsch and Danielle Eisenbud and Dee Cattle and Derek Cheng and Dimitris Paparas and Divyashree Shivakumar Sreepathihalli and Doug Reid and Dustin Tran and Dustin Zelle and Eric Noland and Erwin Huizenga and Eugene Kharitonov and Frederick Liu and Gagik Amirkhanyan and Glenn Cameron and Hadi Hashemi and Hanna Klimczak-Pluci'nska and Harman Singh and Harsh Mehta and Harshal Tushar Lehri and Hussein Hazimeh and Ian Ballantyne and Idan Szpektor and Ivan Nardini and Jean Pouget-Abadie and Jetha Chan and Joe Stanton and J. Michael Wieting and Jonathan Lai and Jordi Orbay and Joe Fernandez and Joshua Newlan and Junsong Ji and Jyotinder Singh and Kat Black and Kathy Yu and Kevin Hui and Kiran Vodrahalli and Klaus Greff and Linhai Qiu and Marcella Valentine and Marina Coelho and Marvin Ritter and Matt Hoffman and Matthew Watson and Mayank Chaturvedi and Michael Moynihan and Min Ma and Nabila Babar and Natasha Noy and Nathan Byrd and Nick Roy and Nikola Momchev and Nilay Chauhan and Oskar Bunyan and Pankil Botarda and Paul Caron and Paul Kishan Rubenstein and Phil Culliton and Philipp Schmid and Pier Giuseppe Sessa and Pingmei Xu and Piotr Stańczyk and Pouya Dehghani Tafti and Rakesh Shivanna and Renjie Wu and Renke Pan and Reza Ardeshir Rokni and Rob Willoughby and Rohith Vallu and Ryan Mullins and Sammy Jerome and Sara Smoot and Sertan Girgin and Shariq Iqbal and Shashir Reddy and Shruti Sheth and Siim P{\~o}der and Sijal Bhatnagar and Sindhu Raghuram Panyam and Sivan Eiger and Susan Zhang and Tianqi Liu and Trevor Yacovone and Tyler Liechty and Uday Kalra and Utku Evci and Vedant Misra and Vincent Roseberry and Vladimir Feinberg and Vlad Kolesnikov and Woohyun Han and Woosuk Kwon and Xi Chen and Yinlam Chow and Yuvein Zhu and Zichuan Wei and Zoltan Egyed and Victor Cotruta and Minh Giang and Phoebe Kirk and Anand Rao and Jessica Lo and Erica Moreira and Luiz Gustavo Martins and Omar Sanseviero and Lucas Gonzalez and Zach Gleicher and Tris Warkentin and Vahab S. Mirrokni and Evan Senter and Eli Collins and Joelle Barral and Zoubin Ghahramani and Raia Hadsell and Yossi Matias and D. Sculley and Slav Petrov and Noah Fiedel and Noam M. Shazeer and Oriol Vinyals and Jeffrey Dean and Demis Hassabis and Koray Kavukcuoglu and Cl{\'e}ment Farabet and Elena Buchatskaya and Jean-Baptiste Alayrac and Rohan Anil and Dmitry Lepikhin and Sebastian Borgeaud and Olivier Bachem and Armand Joulin and Alek Andreev and Cassidy Hardin and Robert Dadashi and L'eonard Hussenot},
  journal={ArXiv},
  year={2025},
  volume={abs/2503.19786},
  url={https://api.semanticscholar.org/CorpusID:277313563}
}

@misc{qwen3technicalreport,
      title={Qwen3 Technical Report}, 
      author={Qwen Team},
      year={2025},
      eprint={2505.09388},
      archivePrefix={arXiv},
      primaryClass={cs.CL},
      url={https://arxiv.org/abs/2505.09388}, 
}

@misc{nussbaum2024nomic,
      title={Nomic Embed: Training a Reproducible Long Context Text Embedder}, 
      author={Zach Nussbaum and John X. Morris and Brandon Duderstadt and Andriy Mulyar},
      year={2024},
      eprint={2402.01613},
      archivePrefix={arXiv},
      primaryClass={cs.CL}
}

@inproceedings{wangdynam3d,
  title={Dynam3D: Dynamic Layered 3D Tokens Empower VLM for Vision-and-Language Navigation},
  author={Wang, Zihan and Lee, Seungjun and Lee, Gim Hee},
  booktitle={The Thirty-ninth Annual Conference on Neural Information Processing Systems},
  year={2025}
}
}


\end{document}